\documentclass[10pt,twocolumn,letterpaper]{article}

\usepackage[letterpaper,text={6.875in,8.875in},top=1in,hmarginratio=1:1,columnsep=0.3125in]{geometry}
\usepackage{times}
\usepackage{xspace}
\usepackage[dvipsnames]{xcolor}
\usepackage{graphicx}
\usepackage{amsmath}
\usepackage{amssymb}
\usepackage{booktabs}
\usepackage[numbers,sort&compress]{natbib}
\usepackage[format=plain,labelformat=simple,labelsep=period,font=small,compatibility=false]{caption}
\usepackage[font=footnotesize,skip=3pt,subrefformat=parens]{subcaption}
\usepackage[hyphens]{url}
\usepackage[shortlabels,inline]{enumitem}
\setlist[itemize]{noitemsep,leftmargin=*,topsep=0em}
\setlist[enumerate]{noitemsep,leftmargin=*,topsep=0em}
\usepackage{needspace}

\newcommand{\method}{UnionCE\xspace}

\definecolor{linkblue}{rgb}{0.21,0.49,0.74}
\usepackage[pagebackref,breaklinks,colorlinks,allcolors=linkblue]{hyperref}

\title{UnionCE: Union-Based Cross-Entropy for 3D Instance-Aware Medical Segmentation}
\author{Luc Bouteille\textsuperscript{1}\quad
Frederic Jonske\textsuperscript{1}\quad
Jens Kleesiek\textsuperscript{1}\quad
Alexander Jaus\textsuperscript{2}\\[0.5em]
\normalsize\textsuperscript{1}Institute for AI in Medicine (IKIM),
University Hospital Essen, Essen, Germany\\
\normalsize\textsuperscript{2}Institute for Anthropomatics and Robotics (IAR),
Karlsruhe Institute of Technology, Karlsruhe, Germany}
\date{}

\begin{document}
\maketitle

\begin{abstract}
Medical image segmentation often requires finding and segmenting multiple lesions
of widely varying sizes. Each lesion can matter for diagnosis or treatment, yet
standard semantic segmentation losses give large lesions more influence than small ones.
Existing instance-aware losses address this imbalance by giving each ground-truth lesion its own loss term, but leave small false detections weakly supervised and tend to gain recall at the cost of precision.
We introduce \method, a novel loss that accounts for both ground-truth and
predicted objects. \method groups the union of ground-truth and predicted
masks into connected components and applies size-normalized cross-entropy
within their Voronoi regions. This gives missed lesions and isolated false
detections their own supervision, while balancing contributions across object
sizes. We evaluate \method on five 3D medical segmentation datasets spanning MRI, CT,
and PET/CT, using a strong U-Net baseline and a modern transformer architecture.
Averaged across datasets and
architectures, \method improves lesion-wise F1 by 4.8 to 8.6 percentage points
over standard and instance-aware losses and is competitive with nnDetection, a self-configuring detection pipeline.
\method adds less than 3\% to training time and is compatible with existing
segmentation pipelines, allowing it to benefit from the broad body
of research on medical image segmentation.
Code is available at \url{https://github.com/TIO-IKIM/UnionCE}.
\end{abstract}
\section{Introduction}
\label{sec:introduction}

\begin{figure*}[t]
\centering
\includegraphics[width=\textwidth]{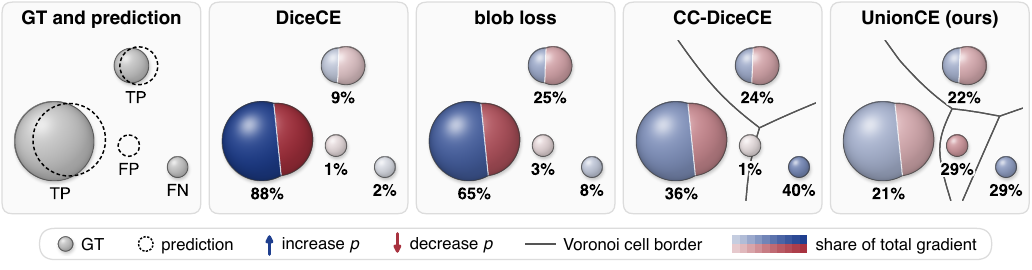}
\caption{Gradient share of the compared losses on a synthetic 3D
scene. Blue surfaces are pushed toward higher foreground
probabilities and red surfaces toward lower ones. Color intensity and
labels indicate each cluster's share of the absolute logit gradient
summed over the displayed components. DiceCE's gradient is dominated by the largest 
cluster. CC-DiceCE is very sensitive to the small FN, but not to the small FP. UnionCE weights clusters more evenly. Lines in CC-DiceCE and UnionCE indicate Voronoi cells.}
\label{fig:headline}
\end{figure*}

In applications where each lesion can affect diagnosis or treatment,
medical image segmentation must reliably identify individual lesions
while capturing their extent and volume. The importance of lesion-level
assessment alongside foreground overlap is well
established~\cite{metricreloaded,ccmetrics,ccdice,blobloss}.
Reliable lesion detection requires both finding true lesions and avoiding
false detections, which can lead to unnecessary follow-up
procedures~\cite{bernstein2023can} and increase clinical review burden.

Detection and instance segmentation often involve complex architectures
and training pipelines~\cite{nndetection,aljabri2022review}.
Lesions in volumetric scans are typically spatially separated, making
semantic segmentation followed by connected-component analysis a natural
way to obtain both accurate masks and individual lesion
predictions~\cite{ccmetrics}.
This approach has become the de facto standard for volumetric lesion segmentation.
In challenges such as AutoPET, BraTS, ISLES, and VALDO,
segmentation networks account for the vast majority of submissions and
are the top-ranked methods, with rankings that also include lesion-level
metrics~\cite{autopet2results,autopet3results,brats23results_intracranial,brats23results_ped,isles22results,valdo}.
Semantic segmentation also benefits from extensive research on
architectures, preprocessing, augmentation, and training, supported by
frameworks such as nnU-Net~\cite{nnunet}.
We therefore pursue ``detection via segmentation'' rather than ``segmentation via detection'': introducing lesion-level supervision through the loss in existing semantic segmentation pipelines.

Prior losses address lesion-size imbalance by reweighting lesions according to their size~\cite{iwdice,iwdiceweighted,iwdicestones,fenneteauiwdice,blobloss,ccdice}
or segmentation difficulty~\cite{hil}.
However, this balancing is based on ground-truth lesions.
Isolated false detections are not normalized by their own size and provide little learning signal for networks.

We introduce \method, which combines size-normalized cross-entropy (CE) with a
partition of the union of ground-truth and predicted masks. We form connected components
of this union and assign each a Voronoi region. Missed lesions, isolated false
positives, and groups of overlapping ground-truth and predicted lesions then
contribute through one common formulation.

Our contributions are:
\begin{enumerate}[label=(\alph*)]
    \item an instance-aware CE objective with union-based regions and explicit
    size normalization for ground-truth and predicted objects;
    \item a theoretical analysis of gradient behavior in existing losses,
    highlighting their limitations for instance-aware supervision;
    \item a controlled comparison against DiceCE, blob loss, and CC-DiceCE
    across five datasets spanning MRI, CT, and PET/CT, using both a U-Net
    and a transformer architecture.
\end{enumerate}

\section{Background and Related Work}
\label{sec:related-work}

\subsection{Notation}
\label{sec:preliminaries}

Let $\Omega\subset\mathbb Z^3$ be a training patch, $y_i\in\{0,1\}$ its
ground-truth label, and $p_i=\operatorname{sigmoid}(z_i)$ the foreground probability.
For two softmax outputs, $z_i$ is the foreground logit minus the background
logit. Write $Y=\{i:y_i=1\}$ for the ground-truth mask and $\widehat Y$
for a hard prediction. Further details on the theoretical analysis are in
Secs.~\ref{sec:supp-global-gradients}--\ref{sec:supp-union-gradients} of the
supplementary material.

\subsection{Base Losses}
\label{sec:base-losses}

The binary cross-entropy (CE) at voxel $i$ is
\begin{equation}
 \ell_i=-y_i\log p_i-(1-y_i)\log(1-p_i).
 \label{eq:voxel-ce}
\end{equation}
For a region $V\subseteq\Omega$, CE and Dice loss are
\begin{equation}
 \begin{aligned}
  \mathcal L_{\mathrm{CE}}(V)
   &=\frac{1}{|V|}\sum_{i\in V}\ell_i,\\
  \mathcal L_{\mathrm{Dice}}(V)
   &=1-\frac{2\sum_{i\in V}p_i y_i+\epsilon}
                 {\sum_{i\in V}p_i+\sum_{i\in V}y_i+\epsilon},
 \end{aligned}
 \label{eq:base-losses}
\end{equation}
where $\epsilon\geq0$ is a smoothing constant. 

\subsection{Volume Bias in Global Losses}
\label{sec:volume-bias}

Let $\{C_j\}_{j=1}^N$ be the 26-connected components of $Y$, with
$s_j=|C_j|$ and $T=|Y|>0$. We compare two controlled cases at the same error
confidence $r\in(1/2,1)$: (a) a missed lesion $C_j$ with $p_i=1-r$ on its
$s_j$ voxels and (b) a false-positive component $F_l\subseteq\Omega\setminus Y$
with $f_l=|F_l|$ and $p_i=r$ on its voxels. In each case, all other predictions are correct.
We first sum the absolute probability gradients over the error region $A\in\{C_j,F_l\}$:
\begin{equation}
 G_{\mathcal L}(A)
 =\sum_{i\in A}
   \left|\frac{\partial\mathcal L}{\partial p_i}\right|.
 \label{eq:component-gradient}
\end{equation}
For global CE and Dice with $\epsilon=0$, these totals are
\begin{equation}
 \begin{aligned}
  G_{\mathrm{CE}}(C_j)&=\frac{s_j}{|\Omega|(1-r)}, &
  G_{\mathrm{Dice}}(C_j)&=\frac{2Ts_j}{(2T-rs_j)^2},\\
  G_{\mathrm{CE}}(F_l)&=\frac{f_l}{|\Omega|(1-r)}, &
  G_{\mathrm{Dice}}(F_l)&=\frac{2Tf_l}{(2T+rf_l)^2}.
 \end{aligned}
 \label{eq:global-component-gradients}
\end{equation}
Neither loss explicitly balances individual components. Small missed lesions and
small isolated false positives can therefore receive little gradient signal,
even though they can be clinically important. Instance-aware losses aim to
address this gradient imbalance.

\subsection{Component-Based Losses}
\label{sec:component-losses}

Let $m(P,Q)$ be a base segmentation loss. Blob loss~\cite{blobloss} computes
a loss for each ground-truth component and averages these losses. In each term,
other ground-truth components are ignored.
\begin{equation}
 B_m(\widehat Y,Y)
 =\frac{1}{N}\sum_{j=1}^N
   m\bigl(\widehat Y\setminus(Y\setminus C_j),C_j\bigr).
 \label{eq:blob-metric}
\end{equation}

CC-DiceCE~\cite{ccdice} instead assigns each voxel to its nearest
ground-truth component, forming Voronoi regions
\begin{equation}
 V_j=\{i\in\Omega:\operatorname{dist}_2(i,C_j)
       \leq\operatorname{dist}_2(i,C_l)\ \forall l\},
 \label{eq:reference-regions}
\end{equation}
where distances are Euclidean and ties are resolved arbitrarily. It averages
the scores within these regions:
\begin{equation}
 \mathrm{CC}_m(\widehat Y,Y)
 =\frac{1}{N}\sum_{j=1}^N m\bigl(\widehat Y\cap V_j,C_j\bigr).
 \label{eq:cc-metric}
\end{equation}

\subsection{Remaining Size Dependence}
\label{sec:remaining-size-dependence}

For the Dice loss versions of CC and blob with $\epsilon=0$, the missed-lesion
case gives
\begin{equation}
 G_{\mathrm{CC}}(C_j)=G_{\mathrm{blob}}(C_j)
 =\frac{2}{N(2-r)^2}.
 \label{eq:reference-component-gradients}
\end{equation}
At fixed confidence, the lesion size cancels exactly. For the false-positive
case, assume $F_l\subseteq V_j\setminus Y$. Then
\begin{equation}
 \begin{aligned}
  G_{\mathrm{CC}}(F_l)&=\frac{f_l}{N}\frac{2s_j}{(2s_j+rf_l)^2},\\
  G_{\mathrm{blob}}(F_l)&=\frac{f_l}{N}\sum_{k=1}^N
    \frac{2s_k}{(2s_k+rf_l)^2}.
 \end{aligned}
 \label{eq:fp-component-gradients}
\end{equation}
The total false-positive gradient still depends on component size for both losses.
For example, in Fig.~\ref{fig:headline}, the small false positive receives
at most 3\% of the total logit gradient for both blob loss and CC-DiceCE. 
Additionally, for blob loss, the false-positive gradient depends on the size of every ground-truth lesion in the patch.
The same size comparisons hold for logit gradients at fixed confidence.

These gradient analyses provide useful guidance for loss design. However,
network training depends on many interacting factors, such as the network
architecture, optimizer, and data sampling, so the findings should be treated
as directional.
\section{Method}
\label{sec:method}

\subsection{Partitioning}
\label{sec:union-partition}

Inspired by the grouping used in Cluster Dice~\cite{clusterdice}, we extract
the 26-connected components $\{C_k\}_{k=1}^K$ of
\begin{equation}
 U=Y\cup\widehat Y.
 \label{eq:union}
\end{equation}
We form Voronoi cells $V_k$ as in Eq.~\ref{eq:reference-regions}, using
the union components. This partitioning naturally handles all three cases:
(a) missed lesions, (b) isolated false positives, and (c) overlapping lesions and predictions (Fig.~\ref{fig:headline}).
We then apply size normalized Cross Entropy to each cell.

\subsection{Why Cross-Entropy?}
\label{sec:why-ce}

For fixed nonnegative weights $a_i$, the weighted CE logit gradient is
\begin{equation}
 \frac{\partial}{\partial z_i}\sum_j a_j\ell_j
 =a_i(p_i-y_i).
 \label{eq:ce-gradient}
\end{equation}
This provides three useful properties for an instance-aware objective.

First, the weights explicitly determine the geometric scaling of the CE gradient.
We can choose the contribution of each region through its size normalizer.
For Dice, this scaling is implicit. Second, Dice's logit gradient contains
the factor $p_i(1-p_i)$:
\begin{equation}
 \frac{\partial\mathcal L_{\mathrm{Dice}}}{\partial z_i}
 =\frac{\partial\mathcal L_{\mathrm{Dice}}}{\partial p_i}
   p_i(1-p_i).
 \label{eq:dice-logit-gradient}
\end{equation}
Unintuitively, this means more confident errors get
less supervision.

Third, CE provides the same scoring rule in cells with and without ground-truth
lesions. In a cell $V$ without ground-truth lesions, smoothed Dice has gradient
\begin{equation}
 \begin{aligned}
  \frac{\partial\mathcal L_{\mathrm{Dice}}(V)}{\partial p_j}
    &=\frac{\epsilon}{(\sum_{i\in V}p_i+\epsilon)^2}.
 \end{aligned}
 \label{eq:empty-dice-gradient}
\end{equation}
For nonzero predicted mass, the gradient vanishes as $\epsilon\to0$; with
positive smoothing, its scale depends on the arbitrary smoothing constant and
the total predicted mass. CE, on the other hand, is naturally symmetric.

These three limitations suggest that, despite its empirical success in medical
image segmentation, Dice loss may be less suited to the problems that
instance-aware losses aim to address.

\subsection{Size Normalization}
\label{sec:union-weighting}

To balance gradient contributions across component sizes, we explicitly normalize
CE within each cell.

Within each cell, let $g_k=|Y\cap V_k|$ and
$h_k=|\widehat Y\cap V_k|$. We normalize isolated false positives and missed
lesions by their own size, and mixed cells by the smaller of $g_k$ and $h_k$:
\begin{equation}
 s_k=
 \begin{cases}
  \min(g_k,h_k), & g_k>0\ \text{and}\ h_k>0,\\
  \max(g_k,h_k), & \text{otherwise}.
 \end{cases}
 \label{eq:cluster-normalizer}
\end{equation}

Using the minimum naturally handles strong under- or oversegmentation.
At fixed confidence, isolated errors ($Y\cap V_k=\varnothing$ or
$\widehat Y\cap V_k=\varnothing$) receive the same total gradient regardless
of size. The total gradient over the whole cell also depends on the
surrounding background probabilities.

\subsection{Complete Objective}
\label{sec:union-objective}

The regional loss is
\begin{equation}
 \mathcal L_{\mathrm{union}}
 =\frac{1}{K}\sum_{k=1}^K\frac{1}{s_k}
   \sum_{i\in V_k}\ell_i.
 \label{eq:union-ce}
\end{equation}

We combine the regional term with ordinary CE:
\begin{equation}
 \mathcal L_{\mathrm{UnionCE}}
 =\underbrace{\frac{1}{|\Omega|}\sum_{i\in\Omega}\ell_i}
   _{\mathcal L_{\mathrm{global}}}
 +\mathcal L_{\mathrm{union}}.
 \label{eq:union-total}
\end{equation}
The global term supervises every voxel uniformly, while the regional term
emphasizes errors relative to union-cluster size. If $U=\varnothing$, the regional
term falls back to global CE.

\subsection{Prediction Mask}
\label{sec:prediction-mask}

Since argmax thresholding is not Dice-optimal for calibrated losses such as
cross-entropy~\cite{nordstrom,rankseg}, we use adaptive top-$k$ selection
following~\cite{nordstrom}. Let $\lambda$ denote the resulting threshold.
The training mask is
\begin{equation}
 \widehat Y=\{i\in\Omega:p_i>\alpha\lambda\},
 \qquad 0<\alpha\leq1.
 \label{eq:training-mask}
\end{equation}
Training computes the threshold per patch, while inference uses the whole image.
Predictions below the training cutoff may therefore enter the final segmentation.
A lower $\alpha$ allows us to supervise these potential false positives.
Inference uses $\alpha = 1$.
\section{Experiments}
\label{sec:experiments}

\subsection{Data}

We evaluate our method on five datasets spanning a wide range of modalities, lesion sizes, and lesion counts
(Table~\ref{tab:data}). CMB is the cerebral microbleed task of VALDO \cite{valdo}.
BraTS-METS is the multisequence brain MRI cohort of the
BraTS 2026 metastases challenges \cite{mets2025,mets2023}. We use tumor core as a binary target and the mandatory T1n, T1c, and
T2f sequences, excluding the partly synthesized optional T2w. For LiTS contrast-enhanced
abdominal CT \cite{lits}, we follow \cite{blobloss} and segment only the liver tumor class to
keep all segmentation targets binary. AutoPET III
combines 1,014 FDG and 597 PSMA PET/CT studies with whole-body tumor lesion annotations
from two clinical centers \cite{autopet1,autopet2}. From ISLES 2026, we use 1,453
labeled native-space T1-weighted MRI scans of stroke lesions, drawn
from multi-center source cohorts \cite{soop,isles2,isles1}.

\begin{table*}[t]
\caption{Cohort characteristics. Empty scans have no ground-truth lesions.
Components use 26-connectivity and native spacing;
component-volume statistics are computed across all components in the dataset. For each case,
intra-case CoV is the standard deviation divided by the mean of its ground-truth component volumes.
Bracketed entries show the first and third quartiles.}
\label{tab:data}
\centering
\small
\setlength{\tabcolsep}{4pt}
\begin{tabular}{@{}lccrrrrcc@{}}
\toprule
& & & \multicolumn{2}{c}{CC/case} & \multicolumn{2}{c}{Component volume (mm$^3$)} & Foreground & Intra-case CoV\\
\cmidrule(lr){4-5}\cmidrule(lr){6-7}
Dataset & Scans & Empty & median & mean & median & mean & (\%) & median\\
\midrule
AutoPET III & 1611 & 573 & 2 [0, 11] & 17.0 & 1294 [495, 3471]   & 8856  & 0.028 & 0.37 [0.00, 1.67]\\
ISLES       & 1453 & 5   & 2 [1, 4]  & 3.4  & 162 [20, 1300]     & 8287  & 0.268 & 0.73 [0.00, 1.28]\\
BraTS-METS  & 1294 & 50  & 4 [1, 8]  & 7.6  & 43.0 [14.5, 185.8] & 742   & 0.056 & 0.90 [0.00, 1.38]\\
LiTS        & 131  & 13  & 3 [1, 9]  & 6.9  & 539 [145, 2733]    & 11249 & 0.124 & 0.85 [0.00, 1.45]\\
CMB         & 72   & 22  & 1 [0, 2]  & 3.3  & 9.9 [7.3, 17.1]    & 18.7  & 0.001 & 0.00 [0.00, 0.03]\\
\bottomrule
\end{tabular}
\end{table*}

\subsection{Experimental Setup}
\label{sec:training}

We evaluate all losses in the nnU-Net \cite{nnunet} framework, which provides
a strong baseline \cite{nnunetrevisited}. To test the effect across network
architectures, we use the default U-Net configuration and the transformer
architecture PrimusV2-S~\cite{primus}.

\medskip\noindent\textbf{Training Setup.}
We evaluate all losses with five-fold cross-validation and 500 training
epochs per fold. For AutoPET III and BraTS-METS, we group folds by patient to prevent leakage.
Because CMB contains only 50 scans with lesions and very few lesions per scan,
we repeat each segmentation experiment on this dataset three times using the
same folds and average each scan's metrics across runs to reduce sensitivity
to training variability.
We compare \method with
DiceCE, blob loss~\cite{blobloss}, and CC-DiceCE~\cite{ccdice}. 
For fairness, we evaluate all variants with fixed settings across all datasets.
Blob loss and
CC-DiceCE combine their regional DiceCE term with global DiceCE at equal
weight. For \method, we set the training threshold factor to $\alpha=0.75$
(Eq.~\ref{eq:training-mask}).
Following~\cite{ccdice}, all Dice terms use $\epsilon=0$ across datasets to avoid convergence issues on CMB.

We use the AdamW configuration from Primus~\cite{primus} for both backbones.
For U-Net, we adapt it to convolutional networks by setting both the initial learning rate and the weight decay to $10^{-3}$. We use 50 epochs of linear warmup
followed by polynomial learning-rate decay.
Because we observed numerical instabilities across all instance losses,
these variants use bfloat16 mixed precision. 

\medskip\noindent\textbf{Detector baseline.}
A key question is whether detection via segmentation can compete with dedicated
object detection. We therefore compare a segmentation model trained with \method
with nnDetection \cite{nndetection}. To compute F1, we need a threshold for the predicted box scores.
Since nnDetection already tunes post-processing parameters on the validation cases,
we also select this threshold on those cases using a linear sweep to maximize nnDetection's mean F1.
We call this nnDetection (oracle) because tuning and evaluation use the same cases,
making it an optimistic reference for the detector.

\medskip\noindent\textbf{Inference.}
Dice-containing objectives use standard logit aggregation and argmax, since argmax is Dice-optimal \cite{nordstrom}.
For \method, we average softmax probabilities to maintain calibration and
apply the threshold from Sec.~\ref{sec:prediction-mask}. Since top-$k$ always segments something, we return an empty mask if
$\max_i p_i\leq\tau$. We adopt the empty-scan threshold $\tau=0.1$ from the
RankSEG-RMA implementation~\cite{rankseg}.

\subsection{Evaluation}
\label{sec:evaluation}

\noindent\textbf{Metrics.}
We assess (a) foreground overlap with global Dice, (b) lesion detection with
lesion-wise precision, recall, and F1, and (c) lesion segmentation with
Cluster Dice (C-DSC)~\cite{clusterdice}. Global Dice emphasizes larger lesions,
while lesion-wise metrics capture whether individual lesions are found and
how many false detections are produced. C-DSC evaluates segmentation quality
at the cluster level.

For lesion-wise precision, recall, and F1, we use maximum-cardinality matching
of connected components at IoU${}>0.1$. All metrics are macro-averaged
over scans. Following~\cite{valdo}, scans without ground-truth lesions
are excluded from these five metrics. To measure false detections on these
scans, we separately report FP/neg, the mean number of predicted components.

\medskip\noindent\textbf{Box F1.}
Since nnDetection predicts bounding boxes, we use box F1 for this comparison.
We replace each segmentation component with its bounding box and
match boxes with IoU${}>0.1$ using the same one-to-one matching and averaging.

\medskip\noindent\textbf{Statistical analysis.}
We use two-sided Wilcoxon signed-rank tests paired by scan, with
Holm-Bonferroni correction per backbone, dataset, and metric.
On CMB, tests use the metrics averaged over runs for each scan.
$\dagger$ marks $p<0.05$.
\section{Results and Discussion}
\label{sec:results}

\subsection{Segmentation Baselines}

\begin{table*}[t]
\begin{minipage}[t]{0.49\textwidth}
\caption{U-Net results.
Tests vs. \method, Holm over three baselines.}
\label{tab:main}
\centering\footnotesize
\setlength{\tabcolsep}{1pt}
\renewcommand{\arraystretch}{1.1}
\begin{tabular*}{\linewidth}{@{\extracolsep{\fill}}lrrrrrr@{}}
\toprule
Method & Dice $\uparrow$ & C-DSC $\uparrow$ & F1 $\uparrow$ & Prec. $\uparrow$ & Rec. $\uparrow$ & FP/neg $\downarrow$\\
\midrule
\multicolumn{7}{l}{\textit{AutoPET III}}\\
DiceCE & 59.3\textsuperscript{$\dagger$} & \underline{35.1}\textsuperscript{$\dagger$} & \underline{55.2}\textsuperscript{$\dagger$} & \underline{52.4}\textsuperscript{$\dagger$} & 67.1\textsuperscript{$\dagger$} & \underline{5.1}\textsuperscript{$\dagger$}\\
blob loss & 57.5\textsuperscript{$\dagger$} & 32.0\textsuperscript{$\dagger$} & 53.0\textsuperscript{$\dagger$} & 47.4\textsuperscript{$\dagger$} & \underline{71.8}\textsuperscript{$\dagger$} & 7.5\textsuperscript{$\dagger$}\\
CC-DiceCE & \underline{60.5} & 29.0\textsuperscript{$\dagger$} & 48.5\textsuperscript{$\dagger$} & 39.7\textsuperscript{$\dagger$} & \textbf{75.5}\textsuperscript{$\dagger$} & 10.6\textsuperscript{$\dagger$}\\
\textbf{\method} & \textbf{61.8} & \textbf{42.1} & \textbf{62.8} & \textbf{63.4} & 69.7 & \textbf{3.2}\\
\midrule
\multicolumn{7}{l}{\textit{ISLES}}\\
DiceCE & \textbf{65.5}\textsuperscript{$\dagger$} & \underline{41.6} & \underline{59.2}\textsuperscript{$\dagger$} & \underline{65.4}\textsuperscript{$\dagger$} & 65.6\textsuperscript{$\dagger$} & 1.4\\
blob loss & 64.8\textsuperscript{$\dagger$} & 39.3\textsuperscript{$\dagger$} & 57.1\textsuperscript{$\dagger$} & 60.7\textsuperscript{$\dagger$} & \underline{66.2}\textsuperscript{$\dagger$} & \textbf{1.0}\\
CC-DiceCE & \underline{65.1}\textsuperscript{$\dagger$} & 21.6\textsuperscript{$\dagger$} & 36.8\textsuperscript{$\dagger$} & 31.2\textsuperscript{$\dagger$} & \textbf{70.6}\textsuperscript{$\dagger$} & \underline{1.2}\\
\textbf{\method} & 64.4 & \textbf{42.8} & \textbf{62.4} & \textbf{72.9} & 64.7 & 2.8\\
\midrule
\multicolumn{7}{l}{\textit{BraTS-METS}}\\
DiceCE & 75.3\textsuperscript{$\dagger$} & \underline{55.7}\textsuperscript{$\dagger$} & 74.1\textsuperscript{$\dagger$} & \underline{80.3}\textsuperscript{$\dagger$} & 74.6\textsuperscript{$\dagger$} & \underline{1.2}\\
blob loss & 73.0\textsuperscript{$\dagger$} & 54.3\textsuperscript{$\dagger$} & \underline{74.2}\textsuperscript{$\dagger$} & 79.2\textsuperscript{$\dagger$} & 75.7\textsuperscript{$\dagger$} & 1.3\\
CC-DiceCE & \underline{75.5}\textsuperscript{$\dagger$} & 50.6\textsuperscript{$\dagger$} & 70.0\textsuperscript{$\dagger$} & 68.5\textsuperscript{$\dagger$} & \textbf{79.6}\textsuperscript{$\dagger$} & \underline{1.2}\\
\textbf{\method} & \textbf{76.6} & \textbf{59.1} & \textbf{77.4} & \textbf{83.5} & \underline{77.1} & \textbf{1.1}\\
\midrule
\multicolumn{7}{l}{\textit{LiTS}}\\
DiceCE & 57.4 & \underline{37.1}\textsuperscript{$\dagger$} & 58.2\textsuperscript{$\dagger$} & \underline{54.9}\textsuperscript{$\dagger$} & 72.7 & \underline{2.8}\textsuperscript{$\dagger$}\\
blob loss & 57.4 & 35.7\textsuperscript{$\dagger$} & \underline{58.5}\textsuperscript{$\dagger$} & 51.7\textsuperscript{$\dagger$} & \underline{78.0}\textsuperscript{$\dagger$} & 4.5\textsuperscript{$\dagger$}\\
CC-DiceCE & \underline{60.2} & 35.2\textsuperscript{$\dagger$} & 57.7\textsuperscript{$\dagger$} & 50.6\textsuperscript{$\dagger$} & \textbf{79.5}\textsuperscript{$\dagger$} & 4.5\textsuperscript{$\dagger$}\\
\textbf{\method} & \textbf{60.7} & \textbf{45.8} & \textbf{69.3} & \textbf{69.4} & 75.4 & \textbf{1.6}\\
\midrule
\multicolumn{7}{l}{\textit{CMB}}\\
DiceCE & 43.9 & 37.6 & 56.5 & 61.1 & \underline{59.2} & 0.9\\
blob loss & 41.7 & 35.1 & 54.8 & 59.3\textsuperscript{$\dagger$} & 57.8 & 1.0\textsuperscript{$\dagger$}\\
CC-DiceCE & \textbf{45.1} & \textbf{39.2} & \textbf{58.6} & \underline{63.3} & \textbf{60.8} & \underline{0.8}\\
\textbf{\method} & \underline{44.0} & \underline{38.8} & \underline{57.6} & \textbf{65.0} & 57.6 & \textbf{0.5}\\
\bottomrule
\end{tabular*}
\end{minipage}\hfill
\begin{minipage}[t]{0.49\textwidth}
\caption{PrimusV2-S results.
Tests vs. \method, Holm over three baselines.}
\label{tab:primus}
\centering\footnotesize
\setlength{\tabcolsep}{1pt}
\renewcommand{\arraystretch}{1.1}
\begin{tabular*}{\linewidth}{@{\extracolsep{\fill}}lrrrrrr@{}}
\toprule
Method & Dice $\uparrow$ & C-DSC $\uparrow$ & F1 $\uparrow$ & Prec. $\uparrow$ & Rec. $\uparrow$ & FP/neg $\downarrow$\\
\midrule
\multicolumn{7}{l}{\textit{AutoPET III}}\\
DiceCE & \underline{59.9}\textsuperscript{$\dagger$} & \underline{35.5}\textsuperscript{$\dagger$} & 55.6\textsuperscript{$\dagger$} & \underline{53.3}\textsuperscript{$\dagger$} & 66.8\textsuperscript{$\dagger$} & \underline{4.4}\textsuperscript{$\dagger$}\\
blob loss & 59.8\textsuperscript{$\dagger$} & \underline{35.5}\textsuperscript{$\dagger$} & \underline{56.3}\textsuperscript{$\dagger$} & 52.6\textsuperscript{$\dagger$} & \underline{70.1}\textsuperscript{$\dagger$} & 5.0\textsuperscript{$\dagger$}\\
CC-DiceCE & \textbf{61.1}\textsuperscript{$\dagger$} & 34.4\textsuperscript{$\dagger$} & 54.7\textsuperscript{$\dagger$} & 48.8\textsuperscript{$\dagger$} & \textbf{72.6}\textsuperscript{$\dagger$} & 4.8\textsuperscript{$\dagger$}\\
\textbf{\method} & 59.2 & \textbf{40.2} & \textbf{61.2} & \textbf{64.8} & 65.0 & \textbf{2.3}\\
\midrule
\multicolumn{7}{l}{\textit{ISLES}}\\
DiceCE & \textbf{64.7}\textsuperscript{$\dagger$} & \textbf{41.9}\textsuperscript{$\dagger$} & \underline{60.3}\textsuperscript{$\dagger$} & \underline{70.9} & 62.3\textsuperscript{$\dagger$} & \underline{1.6}\\
blob loss & 64.4\textsuperscript{$\dagger$} & \underline{40.2} & 58.7\textsuperscript{$\dagger$} & 65.6\textsuperscript{$\dagger$} & \underline{64.2} & \textbf{1.4}\\
CC-DiceCE & \underline{64.5}\textsuperscript{$\dagger$} & 32.2\textsuperscript{$\dagger$} & 51.1\textsuperscript{$\dagger$} & 49.6\textsuperscript{$\dagger$} & \textbf{68.2}\textsuperscript{$\dagger$} & 2.6\\
\textbf{\method} & 62.6 & \textbf{41.9} & \textbf{61.9} & \textbf{72.1} & 64.0 & 2.2\\
\midrule
\multicolumn{7}{l}{\textit{BraTS-METS}}\\
DiceCE & \underline{74.9}\textsuperscript{$\dagger$} & \underline{55.4}\textsuperscript{$\dagger$} & \underline{74.5}\textsuperscript{$\dagger$} & \underline{82.3} & 74.2\textsuperscript{$\dagger$} & \underline{1.7}\\
blob loss & 72.3\textsuperscript{$\dagger$} & 53.6\textsuperscript{$\dagger$} & 74.0\textsuperscript{$\dagger$} & 80.5\textsuperscript{$\dagger$} & 74.9\textsuperscript{$\dagger$} & 1.8\textsuperscript{$\dagger$}\\
CC-DiceCE & \textbf{75.2}\textsuperscript{$\dagger$} & 54.7\textsuperscript{$\dagger$} & 74.3\textsuperscript{$\dagger$} & 76.1\textsuperscript{$\dagger$} & \textbf{79.1}\textsuperscript{$\dagger$} & 2.2\textsuperscript{$\dagger$}\\
\textbf{\method} & 74.7 & \textbf{57.4} & \textbf{76.9} & \textbf{83.2} & \underline{76.5} & \textbf{1.2}\\
\midrule
\multicolumn{7}{l}{\textit{LiTS}}\\
DiceCE & \underline{62.1} & 43.9 & 64.6 & \underline{65.2} & 70.3 & \textbf{1.2}\\
blob loss & 61.9 & 44.1 & 65.9 & 64.7 & \underline{73.8} & 1.8\\
CC-DiceCE & \textbf{64.0}\textsuperscript{$\dagger$} & \underline{44.3} & \underline{66.2} & 63.3\textsuperscript{$\dagger$} & \textbf{75.3} & \underline{1.4}\\
\textbf{\method} & 59.4 & \textbf{44.6} & \textbf{66.4} & \textbf{66.4} & 73.3 & \textbf{1.2}\\
\midrule
\multicolumn{7}{l}{\textit{CMB}}\\
DiceCE & 35.7\textsuperscript{$\dagger$} & 30.2\textsuperscript{$\dagger$} & 46.4\textsuperscript{$\dagger$} & 53.5\textsuperscript{$\dagger$} & 46.3\textsuperscript{$\dagger$} & \textbf{0.6}\\
blob loss & 34.5\textsuperscript{$\dagger$} & 29.0\textsuperscript{$\dagger$} & 46.6\textsuperscript{$\dagger$} & 52.8\textsuperscript{$\dagger$} & 47.2\textsuperscript{$\dagger$} & 0.8\\
CC-DiceCE & \underline{37.2}\textsuperscript{$\dagger$} & \underline{31.1}\textsuperscript{$\dagger$} & \underline{49.4}\textsuperscript{$\dagger$} & \underline{55.6} & \underline{50.4} & \textbf{0.6}\\
\textbf{\method} & \textbf{44.3} & \textbf{37.9} & \textbf{56.9} & \textbf{63.3} & \textbf{57.9} & \underline{0.7}\\
\bottomrule
\end{tabular*}
\end{minipage}
\par\smallskip
\begin{minipage}{\textwidth}
\footnotesize
Scores are reported as percentages, and FP/neg as the mean number of false positives per negative scan.
Bold and underlining mark the best and second-best displayed values per
backbone and dataset, including ties.
\end{minipage}
\end{table*}

\begin{figure}[t]
\centering
\includegraphics[width=\columnwidth]{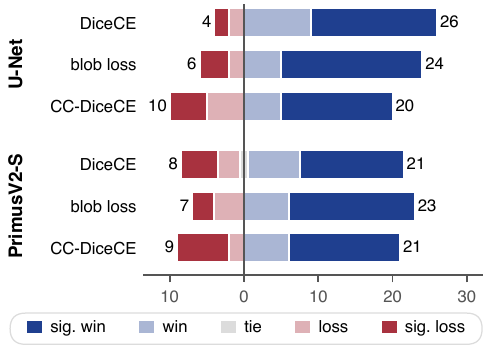}
\caption{Head-to-head comparison of \method with each baseline
for U-Net and PrimusV2-S.
Each bar counts the 30 comparisons (5 datasets $\times$ 6 metrics) that \method wins
(right) or loses (left), based on the mean.
Gray marks exact ties. Dark shades mark significant differences (Sec.~\ref{sec:evaluation}).}
\label{fig:h2h}
\end{figure}

Tables~\ref{tab:main} and~\ref{tab:primus} report the U-Net and PrimusV2-S results.
Across both backbones, \method wins most comparisons against every baseline
(Fig.~\ref{fig:h2h}). Since no single metric captures all
aspects of performance, we discuss lesion detection, lesion segmentation,
and global Dice individually.

\medskip\noindent\textbf{Lesion detection.}
\method achieves the highest mean F1 in nine of ten dataset--architecture
settings, ranking second on CMB with U-Net.
\method's precision exceeds DiceCE
in every setting, by 6.5 percentage points on average, with an even larger
average advantage over CC-DiceCE of 15.7 points. The corresponding mean F1
gains are 4.8 and 8.6 points, respectively. CC-DiceCE, and less
consistently blob loss, tend to have higher recall. Their recall gains
generally come with a larger precision loss, giving \method a better
overall balance. \method also reduces FP/neg on average against all three
baselines.

This pattern is consistent with the gradient analysis. Losses organized around
ground-truth lesions give small false-positive components little supervision relative
to missed lesions. \method normalizes isolated false positives by their
own size, so their gradient contribution does not shrink with component
volume. This provides a possible explanation for the higher precision
and generally lower FP/neg.

\medskip\noindent\textbf{Lesion segmentation.}
\method ranks first in C-DSC in eight settings. Its average gains
over DiceCE, blob loss, and CC-DiceCE are 3.7, 5.2, and 7.8 percentage points,
respectively. Thus, the improved detection balance is accompanied by
better segmentation at the cluster level. 

\medskip\noindent\textbf{Dice.}
Global Dice is dominated by larger lesions and does not capture the lesion-level
detection performance central to this problem~\cite{ccmetrics,ccdice}.
Relative to DiceCE, \method improves global Dice on four datasets with U-Net
but lowers it on four with PrimusV2-S.

\subsection{Detection Baseline}

\begin{table}[t]
\caption{Box metrics for U-Net segmentation models and nnDetection. nnDetection (oracle) has its score threshold tuned on
the evaluated cases (Sec.~\ref{sec:training}). Bold marks the best displayed
value. Tests vs. \method, Holm over two baselines.}
\label{tab:bbox}
\centering\footnotesize
\setlength{\tabcolsep}{2pt}
\renewcommand{\arraystretch}{0.95}
\begin{tabular*}{\columnwidth}{@{\extracolsep{\fill}}lrrrr@{}}
\toprule
Method & Box F1 $\uparrow$ & Box Prec. $\uparrow$ & Box Rec. $\uparrow$ & FP/neg $\downarrow$\\
\midrule
\multicolumn{5}{l}{\textit{AutoPET III}}\\
DiceCE & 55.6\textsuperscript{$\dagger$} & 52.8\textsuperscript{$\dagger$} & 67.6\textsuperscript{$\dagger$} & 5.1\textsuperscript{$\dagger$}\\
nnDetection (oracle) & 51.6\textsuperscript{$\dagger$} & 56.7\textsuperscript{$\dagger$} & 56.7\textsuperscript{$\dagger$} & \textbf{2.9}\textsuperscript{$\dagger$}\\
\textbf{\method} & \textbf{63.0} & \textbf{63.6} & \textbf{69.9} & 3.2\\
\midrule
\multicolumn{5}{l}{\textit{ISLES}}\\
DiceCE & 58.9\textsuperscript{$\dagger$} & 65.0\textsuperscript{$\dagger$} & \textbf{65.1}\textsuperscript{$\dagger$} & 1.4\\
nnDetection (oracle) & 61.9 & \textbf{74.3}\textsuperscript{$\dagger$} & 62.6\textsuperscript{$\dagger$} & \textbf{0.2}\\
\textbf{\method} & \textbf{62.2} & 72.8 & 64.3 & 2.8\\
\midrule
\multicolumn{5}{l}{\textit{BraTS-METS}}\\
DiceCE & 74.4\textsuperscript{$\dagger$} & 80.6\textsuperscript{$\dagger$} & 74.9\textsuperscript{$\dagger$} & 1.2\\
nnDetection (oracle) & 75.7\textsuperscript{$\dagger$} & 83.0 & 74.9\textsuperscript{$\dagger$} & \textbf{0.5}\textsuperscript{$\dagger$}\\
\textbf{\method} & \textbf{77.6} & \textbf{83.7} & \textbf{77.3} & 1.1\\
\midrule
\multicolumn{5}{l}{\textit{LiTS}}\\
DiceCE & 59.6\textsuperscript{$\dagger$} & 56.2\textsuperscript{$\dagger$} & 73.3 & 2.8\textsuperscript{$\dagger$}\\
nnDetection (oracle) & \textbf{70.2} & \textbf{75.1}\textsuperscript{$\dagger$} & 70.9\textsuperscript{$\dagger$} & \textbf{0.6}\textsuperscript{$\dagger$}\\
\textbf{\method} & 69.3 & 69.5 & \textbf{75.4} & 1.6\\
\midrule
\multicolumn{5}{l}{\textit{CMB}}\\
DiceCE & 56.2 & 60.7 & 59.0 & 0.9\\
nnDetection (oracle) & \textbf{59.0} & 59.8 & \textbf{66.1} & 0.8\\
\textbf{\method} & 58.0 & \textbf{65.2} & 58.0 & \textbf{0.5}\\
\bottomrule
\end{tabular*}
\end{table}

\method achieves significantly higher box F1 than nnDetection on AutoPET III and BraTS-METS and is
within about one percentage point on the other three datasets (Table~\ref{tab:bbox}). This comparison uses an
optimistic nnDetection reference, with both post-processing parameters and
score thresholds tuned on the evaluated cases. By comparison, DiceCE has
lower mean box F1 than nnDetection on every dataset except AutoPET III.
These results support detection via segmentation: with lesion-level
supervision, a dense segmentation model can achieve detection performance
comparable to or better than nnDetection while also providing voxel-level
lesion masks. 

\subsection{Ablations}
\label{sec:ablations}

We examine loss and mask design (1--3), CE scoring and region
construction (4--5), and pipeline effects (6--10).
All experiments use U-Net. We run design ablations on BraTS-METS
because the dataset has many scans and a high number of lesion-free cases.
The optimizer, inference, size-filter, and global CE comparisons also cover
all five datasets.

\begin{table}[t]
\caption{Ablations on BraTS-METS with U-Net.
Parenthesized numbers identify the ablations in the text.
Bold marks the best value within each block.
Tests vs. \method use separate Holm corrections for the five design variants
and the three alternative empty-scan thresholds. Tests vs. CC-DiceCE in block 3 are uncorrected.}
\label{tab:ablation}
\centering\footnotesize
\setlength{\tabcolsep}{1pt}
\begin{tabular*}{\columnwidth}{@{\extracolsep{\fill}}lrrrrrr@{}}
\toprule
Variant & Dice $\uparrow$ & C-DSC $\uparrow$ & F1 $\uparrow$ & Prec. $\uparrow$ & Rec. $\uparrow$ & FP/neg $\downarrow$\\
\midrule
\multicolumn{7}{l}{\textit{Loss design}}\\
\textbf{\method} & \textbf{76.6} & \textbf{59.1} & \textbf{77.4} & 83.5 & 77.1 & 1.1\\
(1) Union normalizer & 76.3 & \textbf{59.1} & 76.8\textsuperscript{$\dagger$} & \textbf{85.5}\textsuperscript{$\dagger$} & 74.5\textsuperscript{$\dagger$} & 0.6\textsuperscript{$\dagger$}\\
(2) No threshold scaling & 76.3 & 58.7 & 76.8\textsuperscript{$\dagger$} & 81.9\textsuperscript{$\dagger$} & 77.2 & 1.1\\
(4) DiceCE cells, $\epsilon=0$ & 75.7\textsuperscript{$\dagger$} & 47.0\textsuperscript{$\dagger$} & 67.4\textsuperscript{$\dagger$} & 63.6\textsuperscript{$\dagger$} & \textbf{80.9}\textsuperscript{$\dagger$} & 1.3\\
(4) DiceCE cells, $\epsilon=10$ & 75.0\textsuperscript{$\dagger$} & 57.4\textsuperscript{$\dagger$} & 75.4\textsuperscript{$\dagger$} & 82.7 & 74.2\textsuperscript{$\dagger$} & \textbf{0.5}\textsuperscript{$\dagger$}\\
(6) SGD & 73.5\textsuperscript{$\dagger$} & 55.4\textsuperscript{$\dagger$} & 74.1\textsuperscript{$\dagger$} & 79.0\textsuperscript{$\dagger$} & 75.6\textsuperscript{$\dagger$} & 1.6\\
\midrule
\multicolumn{7}{l}{\textit{(3) Empty-scan threshold $\tau$}}\\
$\tau=0.05$ & \textbf{76.6} & \textbf{59.1} & \textbf{77.5} & \textbf{83.7} & \textbf{77.2} & 1.4\textsuperscript{$\dagger$}\\
$\tau=0.1$ (\method) & \textbf{76.6} & \textbf{59.1} & 77.4 & 83.5 & 77.1 & 1.1\\
$\tau=0.3$ & 76.4 & 59.0 & 77.2 & 83.2 & 76.8 & \textbf{0.6}\textsuperscript{$\dagger$}\\
$\tau=0.5$ & 76.1 & 58.7 & 76.7\textsuperscript{$\dagger$} & 82.7\textsuperscript{$\dagger$} & 76.4\textsuperscript{$\dagger$} & \textbf{0.6}\textsuperscript{$\dagger$}\\
\midrule
\multicolumn{7}{l}{\textit{(5) Ground-truth cells}}\\
CC-DiceCE & 75.5 & 50.6 & 70.0 & 68.5 & \textbf{79.6} & \textbf{1.2}\\
CC-CE & \textbf{76.6}\textsuperscript{$\dagger$} & \textbf{56.3}\textsuperscript{$\dagger$} & \textbf{75.0}\textsuperscript{$\dagger$} & \textbf{77.1}\textsuperscript{$\dagger$} & 79.1\textsuperscript{$\dagger$} & 1.9\textsuperscript{$\dagger$}\\
\bottomrule
\end{tabular*}
\end{table}

\noindent\textbf{(1) Mixed-cell weighting.}
To assess the effect of upweighting cells containing both ground-truth and
predicted foreground ($g_k>0$ and $h_k>0$),
we replace the minimum normalizer in Eq.~\ref{eq:cluster-normalizer}
with the union-cluster size, $|C_k|$.
This downweights mixed cells and leaves the weights of isolated false positives
and missed lesions unchanged. The weaker supervision of poorly segmented
lesions shifts the balance toward higher precision
and lower recall (Table~\ref{tab:ablation}, ablation~1).

\medskip\noindent\textbf{(2) Training-mask threshold.}
To assess whether exposing additional potential
false positives helps, we disable threshold scaling by setting $\alpha=1$
in Eq.~\ref{eq:training-mask}.
Without scaling, precision drops while recall remains nearly unchanged
(Table~\ref{tab:ablation}, ablation~2), consistent with the benefit of supervising
potential false positives below the patch-level cutoff.

\medskip\noindent\textbf{(3) Empty-scan threshold.}
To assess sensitivity to the empty-scan threshold, we evaluate the same trained checkpoints at
$\tau\in\{0.05,0.1,0.3,0.5\}$.
Lowering $\tau$ from 0.1 to 0.05 admits predictions on a few more scans,
slightly improving lesion-positive metrics but significantly increasing
FP/neg from 1.1 to 1.4. Higher thresholds slightly lower these metrics
while significantly reducing FP/neg to 0.6 (Table~\ref{tab:ablation}, ablation~3).
The threshold therefore controls a trade-off between retaining low-confidence
predictions and suppressing false positives on lesion-free scans.

\medskip\noindent\textbf{(4) DiceCE in union cells.}
To test the derived limitations of Dice
in union cells (Sec.~\ref{sec:why-ce}), we keep the union partition and
training mask but score each cell with DiceCE instead of CE, using argmax decoding.
In cells without ground-truth lesions, we compare $\epsilon=0$ with
$\epsilon=10$, since small smoothing values give false-positive cells
little to no Dice gradient (Eq.~\ref{eq:empty-dice-gradient}).
Both settings lower F1 (Table~\ref{tab:ablation}, ablation~4).

With $\epsilon=0$, the Dice term gives isolated false positives no gradient.
CE is averaged over the cell, so its total gradient scales with the
false-positive component's size relative to the cell. Precision drops sharply,
while recall rises above CC-DiceCE, matching the imbalance predicted in
Sec.~\ref{sec:why-ce}.

Setting $\epsilon=10$ restores the missing Dice gradient: precision recovers
substantially, while recall drops. Both remain below the full \method
configuration. We attribute this incomplete
recovery to the size dependence of the smoothed Dice gradient. For a
false-positive component of $n$ voxels at confidence $p$, with negligible
predicted mass elsewhere in the cell, the total pull is
\begin{equation}
 G^z_{\mathrm{Dice}}
 =\frac{n\epsilon p(1-p)}{(np+\epsilon)^2}.
 \label{eq:smoothed-fp-pull}
\end{equation}
At fixed confidence, this peaks when $np=\epsilon$ and decreases for both
smaller and larger components. Smoothing therefore favors a particular
false-positive scale, whereas \method gives each isolated false-positive
component a size-independent total gradient.

\medskip\noindent\textbf{(5) Ground-truth cells.}
To separate the effects of CE scoring and union-based regions,
we use cells seeded only by ground-truth
components, each normalized by its component's size. This gives a CE
counterpart of CC-DiceCE, which we call CC-CE.
Compared with CC-DiceCE, CC-CE improves F1 from 70.0 to 75.0,
with nearly matched recall and higher precision (Table~\ref{tab:ablation}, ablation~5).
We theorize that the relative weighting of background and foreground errors
could explain this. For equally confident errors within the same cell,
\begin{equation}
 \frac{|\partial\mathcal L_{\mathrm{Dice}}/\partial z_{\mathrm{bg}}|}
      {|\partial\mathcal L_{\mathrm{Dice}}/\partial z_{\mathrm{fg}}|}
 =\frac{d}{2-d}\leq1,
 \label{eq:gt-gradient-balance}
\end{equation}
where $d$ is the cell's soft Dice score.
Dice therefore gives background errors less relative weight, especially
at low overlap, while CE treats these errors symmetrically.

CC-CE remains below \method in F1 and
precision and produces more false positives on negative scans than either
\method or CC-DiceCE.

\medskip\noindent\textbf{(6) Optimizer control.}
To control for the training recipe, we compare our AdamW setup with nnU-Net's
SGD recipe for all three baselines on all five datasets and for \method on
BraTS-METS. All three baselines achieve higher F1 with AdamW on most datasets
(Table~\ref{tab:sgd} in the supplementary material). \method shows the same
pattern on BraTS-METS, where switching to SGD lowers F1, precision, and recall
(Table~\ref{tab:ablation}, ablation~6).

We therefore recommend AdamW for \method. Its adaptive updates may help handle
the large gradient norms observed with size-normalized CE. With our training
setup, we observed no non-finite steps across all runs.

\medskip\noindent\textbf{(7) Inference control.}
We swap the full inference setups between \method and the baselines, keeping
the trained models fixed. The baselines receive probability averaging,
adaptive top-$k$ selection, and the same empty-scan threshold $\tau=0.1$.
\method receives logit averaging and argmax, without the empty-scan threshold.

The top-$k$ inference setup improves \method's F1 on all five datasets but generally
lowers F1 for the other losses, with CMB being the main exception (Table~\ref{tab:inference} in the supplementary material). This agrees with
prior work showing that adaptive thresholding is important for CE-based
segmentation~\cite{nordstrom,rankseg}.

\medskip\noindent\textbf{(8) Threshold tuning.}
To test whether threshold tuning can recover \method's gains, we sweep
the logit threshold from $-8$ to $8$ in steps of $0.5$ for DiceCE, blob loss,
and CC-DiceCE on BraTS-METS, using the same trained models.
For foreground probability $p_i=\sigma(z_i)$,
$z_i>\delta\Longleftrightarrow p_i>\sigma(\delta)$, so positive shifts
raise the probability threshold, while $\delta=0$ gives
standard argmax decoding.

\begin{figure}[t]
\centering
\includegraphics[width=\columnwidth]{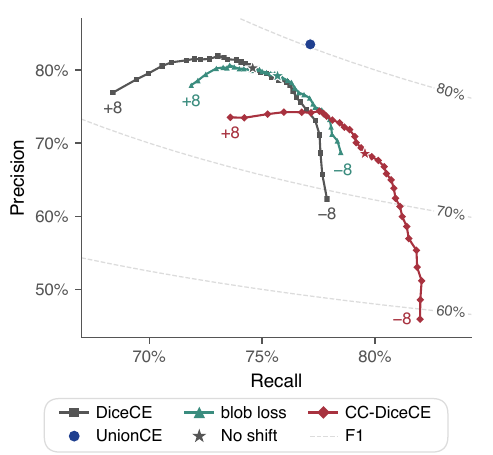}
\caption{Threshold tuning on BraTS-METS with U-Net (ablation~8).
Dashed lines show F1 computed from the plotted mean precision and recall.}
\label{fig:shift-pr}
\end{figure}

Threshold tuning mainly trades precision for recall. It improves F1 for
CC-DiceCE but changes it little for DiceCE and blob loss. Extreme shifts lower
F1 for all three baselines. All three remain below \method, and no tested
threshold matches both its precision and recall
(Fig.~\ref{fig:shift-pr}).

\begin{figure}[t]
\centering
\includegraphics[width=\columnwidth]{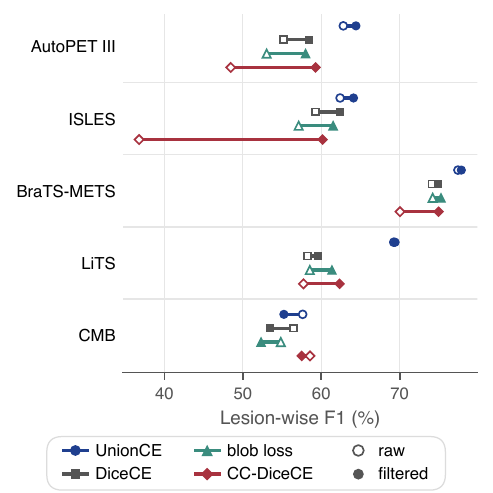}
\caption{Lesion-wise F1 before and after minimum-volume filtering with
U-Net (ablation~9). For each loss and dataset, thresholds are tuned on one fold and
evaluated on the remaining four.}
\label{fig:size-filter}
\end{figure}

\medskip\noindent\textbf{(9) Size filter.}
To test whether removing small false detections can recover \method's gains,
we evaluate a post-processing size filter for all four U-Net losses.
For each loss and dataset, we select the volume
threshold that maximizes mean lesion-wise F1 on one fold and evaluate it on
the remaining four. We repeat this procedure with each fold and report the
average performance.

Removing small predicted components substantially improves the baselines' F1 (Fig.~\ref{fig:size-filter}),
particularly for CC-DiceCE, since it predicts many small components, as
theorized in~\cite{ccdice}. But after filtering, CC-DiceCE loses its recall
advantage over unfiltered \method on four of five datasets, with still
lower precision. Without filtering, \method achieves higher
mean F1 than the filtered baselines in all 15 comparisons. In general, we
recommend against filtering, as it (a) is brittle, (b) depends on the exact
cohort distribution, and (c) comes close to metric gaming by discarding all
predictions below a fixed size.

\medskip\noindent\textbf{(10) Global CE.}
To separate the effect of the pipeline from that of the regional loss,
we remove \method's regional term while keeping its training recipe and
inference. We compare this global CE variant with DiceCE.

Global CE improves F1 over DiceCE on BraTS-METS and CMB but lowers it on
the other three datasets and increases FP/neg on all five
(Table~\ref{tab:ce} in the supplementary material). The CE pipeline alone therefore does not explain
\method's gains.

\subsection{Limitations and Future Work}
\label{sec:limitations}

Our findings also point to several directions for future work.
(a) Our gradient analysis explains component weighting under controlled
conditions, but does not describe the full training dynamics.
(b) Size normalization gives small components substantial weight. Its effect
on sensitivity to annotation noise remains to be studied.
(c) We evaluate only binary lesion segmentation. Extending \method to
multiclass segmentation could follow the one-versus-rest approach of~\cite{kundu}.
\section{Conclusion}
\label{sec:conclusion}

We introduced \method, an instance-aware loss for medical image segmentation that
improves lesion-wise F1 over all baselines in most configurations
(Tables~\ref{tab:main} and~\ref{tab:primus}) and is competitive with an optimistic
detection baseline (Table~\ref{tab:bbox}). We evaluated it across convolutional and
transformer architectures and tested its design choices through ablations
(Table~\ref{tab:ablation}).

{
    \small
    \bibliographystyle{ieeenat_fullname}
    \bibliography{UnionCE}

\begin{thebibliography}{33}
\providecommand{\natexlab}[1]{#1}
\providecommand{\url}[1]{\texttt{#1}}
\expandafter\ifx\csname urlstyle\endcsname\relax
  \providecommand{\doi}[1]{doi: #1}\else
  \providecommand{\doi}{doi: \begingroup \urlstyle{rm}\Url}\fi

\bibitem[Absher et~al.(2024)Absher, Goncher, Newman-Norlund, Perkins,
  Yourganov, Vargas, Sivakumar, Parti, Sternberg, Teghipco, et~al.]{soop}
John Absher, Sarah Goncher, Roger Newman-Norlund, Nicholas Perkins, Grigori
  Yourganov, Jan Vargas, Sanjeev Sivakumar, Naveen Parti, Shannon Sternberg,
  Alex Teghipco, et~al.
\newblock The stroke outcome optimization project: Acute ischemic strokes from
  a comprehensive stroke center.
\newblock \emph{Scientific Data}, 11\penalty0 (1):\penalty0 839, 2024.

\bibitem[Aljabri and AlGhamdi(2022)]{aljabri2022review}
Manar Aljabri and Manal AlGhamdi.
\newblock A review on the use of deep learning for medical images segmentation.
\newblock \emph{Neurocomputing}, 506:\penalty0 311--335, 2022.

\bibitem[Baumgartner et~al.(2021)Baumgartner, J{\"a}ger, Isensee, and
  Maier-Hein]{nndetection}
Michael Baumgartner, Paul~F. J{\"a}ger, Fabian Isensee, and Klaus~H.
  Maier-Hein.
\newblock {nnDetection}: A self-configuring method for medical object
  detection.
\newblock In \emph{Medical Image Computing and Computer Assisted Intervention},
  pages 530--539, 2021.

\bibitem[Bernstein et~al.(2023)Bernstein, Atalay, Dibble, Maxwell, Karam,
  Agarwal, Ward, Healey, and Baird]{bernstein2023can}
Michael~H Bernstein, Michael~K Atalay, Elizabeth~H Dibble, Aaron~WP Maxwell,
  Adib~R Karam, Saurabh Agarwal, Robert~C Ward, Terrance~T Healey, and
  Grayson~L Baird.
\newblock Can incorrect artificial intelligence (ai) results impact
  radiologists, and if so, what can we do about it? a multi-reader pilot study
  of lung cancer detection with chest radiography.
\newblock \emph{European radiology}, 33\penalty0 (11):\penalty0 8263--8269,
  2023.

\bibitem[Bilic et~al.(2023)Bilic, Christ, Li, Vorontsov, Ben-Cohen, Kaissis,
  Szeskin, Jacobs, Mamani, Chartrand, et~al.]{lits}
Patrick Bilic, Patrick Christ, Hongwei~Bran Li, Eugene Vorontsov, Avi
  Ben-Cohen, Georgios Kaissis, Adi Szeskin, Colin Jacobs, Gabriel
  Efrain~Humpire Mamani, Gabriel Chartrand, et~al.
\newblock The liver tumor segmentation benchmark ({LiTS}).
\newblock \emph{Medical Image Analysis}, 84:\penalty0 102680, 2023.

\bibitem[Bouteille et~al.(2026)Bouteille, Jaus, Kleesiek, Stiefelhagen, and
  Heine]{ccdice}
Luc Bouteille, Alexander Jaus, Jens Kleesiek, Rainer Stiefelhagen, and Lukas
  Heine.
\newblock Learning to look closer: A new instance-wise loss for small cerebral
  lesion segmentation.
\newblock In \emph{IEEE International Symposium on Biomedical Imaging}, pages
  1--5, 2026.

\bibitem[De~La~Rosa et~al.(2025)De~La~Rosa, Reyes, Liew, Hutton, Wiest,
  Kaesmacher, Hanning, Hakim, Zubal, Valenzuela, et~al.]{isles22results}
Ezequiel De~La~Rosa, Mauricio Reyes, Sook-Lei Liew, Alexandre Hutton, Roland
  Wiest, Johannes Kaesmacher, Uta Hanning, Arsany Hakim, Richard Zubal, Waldo
  Valenzuela, et~al.
\newblock Deepisles: a clinically validated ischemic stroke segmentation model
  from the isles'22 challenge.
\newblock \emph{Nature communications}, 16\penalty0 (1):\penalty0 7357, 2025.

\bibitem[Dexl et~al.(2026{\natexlab{a}})Dexl, Gatidis, Fr{\"u}h, Jeblick,
  Mittermeier, St{\"u}ber, Schachtner, Topalis, Fabritius, Gu,
  et~al.]{autopet2results}
Jakob Dexl, Sergios Gatidis, Marcel Fr{\"u}h, Katharina Jeblick, Andreas
  Mittermeier, Anna~Theresa St{\"u}ber, Balthasar Schachtner, Johanna Topalis,
  Matthias~P Fabritius, Sijing Gu, et~al.
\newblock Autopet challenge on fully automated lesion segmentation in oncologic
  pet/ct imaging, part 2: domain generalization.
\newblock \emph{Journal of Nuclear Medicine}, 67\penalty0 (3):\penalty0
  481--488, 2026{\natexlab{a}}.

\bibitem[Dexl et~al.(2026{\natexlab{b}})Dexl, Jeblick, Mittermeier, Schachtner,
  St{\"u}ber, Topalis, Rokuss, Isensee, Maier-Hein, Kalisch,
  et~al.]{autopet3results}
Jakob Dexl, Katharina Jeblick, Andreas Mittermeier, Balthasar Schachtner,
  Anna~Theresa St{\"u}ber, Johanna Topalis, Maximilian Rokuss, Fabian Isensee,
  Klaus~H Maier-Hein, Hamza Kalisch, et~al.
\newblock The autopet3 challenge: Automated lesion segmentation in whole-body
  pet/ct multitracer multicenter generalization.
\newblock \emph{arXiv preprint arXiv:2605.05775}, 2026{\natexlab{b}}.

\bibitem[Fathi~Kazerooni et~al.(2025)Fathi~Kazerooni, Khalili, Liu, Haldar,
  Jiang, Zapaishchykova, Pavaine, Shah, Jones, Sheth,
  et~al.]{brats23results_ped}
Anahita Fathi~Kazerooni, Nastaran Khalili, Xinyang Liu, Debanjan Haldar, Zhifan
  Jiang, Anna Zapaishchykova, Julija Pavaine, Lubdha~M. Shah, Blaise~V. Jones,
  Nakul Sheth, et~al.
\newblock Brats-peds: Results of the multi-consortium international pediatric
  brain tumor segmentation challenge 2023.
\newblock \emph{Machine Learning for Biomedical Imaging}, 2025:\penalty0
  72--87, 2025.

\bibitem[Fenneteau et~al.(2022)Fenneteau, Helbert, Bourdon, M'Rabet,
  Fernandez-Maloigne, and Guillevin]{fenneteauiwdice}
Alexandre Fenneteau, David Helbert, Pascal Bourdon, Imane M'Rabet, Christine
  Fernandez-Maloigne, and R{\'e}my Guillevin.
\newblock A size-adaptative segmentation method for better detection of
  multiple sclerosis lesions.
\newblock Preprint, HAL: hal-03836787v1, 2022.

\bibitem[Gatidis et~al.(2022)Gatidis, Hepp, Fr{\"u}h, La~Foug{\`e}re, Nikolaou,
  Pfannenberg, Sch{\"o}lkopf, K{\"u}stner, Cyran, and Rubin]{autopet1}
Sergios Gatidis, Tobias Hepp, Marcel Fr{\"u}h, Christian La~Foug{\`e}re,
  Konstantin Nikolaou, Christina Pfannenberg, Bernhard Sch{\"o}lkopf, Thomas
  K{\"u}stner, Clemens Cyran, and Daniel Rubin.
\newblock A whole-body fdg-pet/ct dataset with manually annotated tumor
  lesions.
\newblock \emph{Scientific Data}, 9\penalty0 (1):\penalty0 601, 2022.

\bibitem[Isensee et~al.(2021)Isensee, Jaeger, Kohl, Petersen, and
  Maier-Hein]{nnunet}
Fabian Isensee, Paul~F. Jaeger, Simon A.~A. Kohl, Jens Petersen, and Klaus~H.
  Maier-Hein.
\newblock {nnU-Net}: A self-configuring method for deep learning-based
  biomedical image segmentation.
\newblock \emph{Nature Methods}, 18\penalty0 (2):\penalty0 203--211, 2021.

\bibitem[Isensee et~al.(2024)Isensee, Wald, Ulrich, Baumgartner, Roy,
  Maier-Hein, and Jaeger]{nnunetrevisited}
Fabian Isensee, Tassilo Wald, Constantin Ulrich, Michael Baumgartner, Saikat
  Roy, Klaus Maier-Hein, and Paul~F Jaeger.
\newblock nnu-net revisited: A call for rigorous validation in 3d medical image
  segmentation.
\newblock In \emph{International conference on medical image computing and
  computer-assisted intervention}, pages 488--498. Springer, 2024.

\bibitem[Jaus et~al.(2025)Jaus, Seibold, Rei{\ss}, Marinov, Li, Ye, Krieg,
  Kleesiek, and Stiefelhagen]{ccmetrics}
Alexander Jaus, Constantin~Marc Seibold, Simon Rei{\ss}, Zdravko Marinov, Keyi
  Li, Zeling Ye, Stefan Krieg, Jens Kleesiek, and Rainer Stiefelhagen.
\newblock Every component counts: Rethinking the measure of success for medical
  semantic segmentation in multi-instance segmentation tasks.
\newblock In \emph{Proceedings of the AAAI Conference on Artificial
  Intelligence}, pages 3904--3912, 2025.

\bibitem[Jeblick et~al.(2024)]{autopet2}
Katharina Jeblick et~al.
\newblock A whole-body psma-pet/ct dataset with manually annotated tumor
  lesions (psma-pet-ct-lesions), 2024.

\bibitem[Jiang et~al.(2024)Jiang, Li, Yi, Chen, and Wang]{hil}
Weili Jiang, Yiming Li, Zhang Yi, Mao Chen, and Jianyong Wang.
\newblock Multi-instance imbalance semantic segmentation by instance-dependent
  attention and adaptive hard instance mining.
\newblock \emph{Knowledge-Based Systems}, 304:\penalty0 112554, 2024.

\bibitem[Kofler et~al.(2023)Kofler, Shit, Ezhov, Fidon, Horvath, Al-Maskari,
  Li, Bhatia, Loehr, Piraud, et~al.]{blobloss}
Florian Kofler, Suprosanna Shit, Ivan Ezhov, Lucas Fidon, Izabela Horvath, Rami
  Al-Maskari, Hongwei~Bran Li, Harsharan Bhatia, Timo Loehr, Marie Piraud,
  et~al.
\newblock blob loss: Instance imbalance aware loss functions for semantic
  segmentation.
\newblock In \emph{Information Processing in Medical Imaging}, pages 755--767,
  2023.

\bibitem[Kundu et~al.(2025)Kundu, Kujawa, Ivory, Barfoot, Shapey, and
  Vercauteren]{clusterdice}
Soumya~Snigdha Kundu, Aaron Kujawa, Marina Ivory, Theodore Barfoot, Jonathan
  Shapey, and Tom Vercauteren.
\newblock {Cluster Dice}: A simple and fast approach for instance-based
  semantic segmentation evaluation via many-to-many matching.
\newblock In \emph{Medical Imaging 2025: Computer-Aided Diagnosis}, page
  1340711. SPIE, 2025.

\bibitem[Kundu et~al.(2026)Kundu, Kofler, Ivory, M{\"o}ller, Shapey, and
  Vercauteren]{kundu}
Soumya~Snigdha Kundu, Florian Kofler, Marina Ivory, Hendrik M{\"o}ller,
  Jonathan Shapey, and Tom Vercauteren.
\newblock Instance awareness of multi-class semantic segmentation loss
  functions.
\newblock In \emph{Proceedings of the IEEE/CVF Conference on Computer Vision
  and Pattern Recognition (CVPR) Workshops}, pages 6680--6688, 2026.

\bibitem[LaBella et~al.(2025)LaBella, Baid, Khanna, McBurney-Lin, McLean,
  Nedelec, Rashid, Tahon, Altes, Bhalerao, et~al.]{brats23results_intracranial}
Dominic LaBella, Ujjwal Baid, Omaditya Khanna, Shan McBurney-Lin, Ryan McLean,
  Pierre Nedelec, Arif Rashid, Nourel~Hoda Tahon, Talissa Altes, Radhika
  Bhalerao, et~al.
\newblock Analysis of the brats 2023 intracranial meningioma segmentation
  challenge.
\newblock \emph{Machine Learning for Biomedical Imaging}, 2025:\penalty0
  38--58, 2025.

\bibitem[Liew et~al.(2018)Liew, Anglin, Banks, Sondag, Ito, Kim, Chan, Ito,
  Jung, Khoshab, et~al.]{isles2}
Sook-Lei Liew, Julia~M Anglin, Nick~W Banks, Matt Sondag, Kaori~L Ito, Hosung
  Kim, Jennifer Chan, Joyce Ito, Connie Jung, Nima Khoshab, et~al.
\newblock A large, open source dataset of stroke anatomical brain images and
  manual lesion segmentations.
\newblock \emph{Scientific data}, 5\penalty0 (1):\penalty0 180011, 2018.

\bibitem[Liew et~al.(2022)Liew, Tavenner, Donnelly, Zavaliangos-Petropulu,
  Jeong, Barisano, Hutton, Simon, Juliano, Suri, et~al.]{isles1}
Sook-Lei Liew, Bethany~P Tavenner, Miranda~R Donnelly, Artemis
  Zavaliangos-Petropulu, Jessica~N Jeong, Giuseppe Barisano, Alexandre Hutton,
  Julia~P Simon, Julia~M Juliano, Anisha Suri, et~al.
\newblock A large, curated, open-source stroke neuroimaging dataset to improve
  lesion segmentation algorithms.
\newblock \emph{Scientific data}, 9\penalty0 (1):\penalty0 320, 2022.

\bibitem[Maier-Hein et~al.(2024)Maier-Hein, Reinke, Godau, Tizabi, B{\"u}ttner,
  Christodoulou, Glocker, Isensee, Kleesiek, Kozubek, et~al.]{metricreloaded}
Lena Maier-Hein, Annika Reinke, Patrick Godau, Minu~D. Tizabi, Florian
  B{\"u}ttner, Evangelia Christodoulou, Ben Glocker, Fabian Isensee, Jens
  Kleesiek, Michal Kozubek, et~al.
\newblock Metrics reloaded: Recommendations for image analysis validation.
\newblock \emph{Nature Methods}, 21\penalty0 (2):\penalty0 195--212, 2024.

\bibitem[Maleki et~al.(2025)Maleki, Amiruddin, Moawad, Yordanov, Gkampenis,
  Fehringer, Umeh, Chukwurah, Memon, Petrovic, et~al.]{mets2025}
Nazanin Maleki, Raisa Amiruddin, Ahmed~W Moawad, Nikolay Yordanov, Athanasios
  Gkampenis, Pascal Fehringer, Fabian Umeh, Crystal Chukwurah, Fatima Memon,
  Bojan Petrovic, et~al.
\newblock Analysis of the miccai brain tumor segmentation--metastases
  (brats-mets) 2025 lighthouse challenge: brain metastasis segmentation on
  pre-and post-treatment mri.
\newblock \emph{arXiv preprint arXiv:2504.12527}, 2025.

\bibitem[Moawad et~al.(2024)Moawad, Janas, Baid, Ramakrishnan, Saluja, Ashraf,
  Maleki, Jekel, Yordanov, Fehringer, et~al.]{mets2023}
Ahmed~W Moawad, Anastasia Janas, Ujjwal Baid, Divya Ramakrishnan, Rachit
  Saluja, Nader Ashraf, Nazanin Maleki, Leon Jekel, Nikolay Yordanov, Pascal
  Fehringer, et~al.
\newblock The brain tumor segmentation ({BraTS-METS}) challenge 2023: Brain
  metastasis segmentation on pre-treatment {MRI}.
\newblock \emph{arXiv preprint arXiv:2306.00838}, 2024.

\bibitem[Nichyporuk et~al.(2021)Nichyporuk, Szeto, Arnold, and
  Arbel]{iwdiceweighted}
Brennan Nichyporuk, Justin Szeto, Douglas~L Arnold, and Tal Arbel.
\newblock Optimizing operating points for high performance lesion detection and
  segmentation using lesion size reweighting.
\newblock \emph{arXiv preprint arXiv:2107.12978}, 2021.

\bibitem[Nordstr{\"o}m et~al.(2025)Nordstr{\"o}m, Maki, and Hult]{nordstrom}
Marcus Nordstr{\"o}m, Atsuto Maki, and Henrik Hult.
\newblock The impact label noise and choice of threshold has on cross-entropy
  and soft-dice in image segmentation.
\newblock In \emph{Proceedings of the IEEE/CVF Conference on Computer Vision
  and Pattern Recognition}, pages 20820--20829, 2025.

\bibitem[Preedanan et~al.(2023)Preedanan, Suzuki, Kondo, Kobayashi, Tanaka,
  Ishioka, Matsuoka, Fujii, and Kumazawa]{iwdicestones}
Wongsakorn Preedanan, Kenji Suzuki, Toshiaki Kondo, Masaki Kobayashi, Hajime
  Tanaka, Junichiro Ishioka, Yoh Matsuoka, Yasuhisa Fujii, and Itsuo Kumazawa.
\newblock Urinary stones segmentation in abdominal {X}-ray images using
  cascaded {U}-net pipeline with stone-embedding augmentation and lesion-size
  reweighting approach.
\newblock \emph{IEEE Access}, 11:\penalty0 25702--25712, 2023.

\bibitem[Shirokikh et~al.(2020)Shirokikh, Shevtsov, Kurmukov, Dalechina,
  Krivov, Kostjuchenko, Golanov, and Belyaev]{iwdice}
Boris Shirokikh, Alexey Shevtsov, Anvar Kurmukov, Alexandra Dalechina, Egor
  Krivov, Valery Kostjuchenko, Andrey Golanov, and Mikhail Belyaev.
\newblock Universal loss reweighting to balance lesion size inequality in {3D}
  medical image segmentation.
\newblock In \emph{Medical Image Computing and Computer Assisted Intervention},
  pages 523--532. Springer, 2020.

\bibitem[Sudre et~al.(2024)Sudre, Van~Wijnen, Dubost, Adams, Atkinson, Barkhof,
  Birhanu, Bron, Camarasa, Chaturvedi, et~al.]{valdo}
Carole~H. Sudre, Kimberlin Van~Wijnen, Florian Dubost, Hieab Adams, David
  Atkinson, Frederik Barkhof, Mahlet~A. Birhanu, Esther~E. Bron, Robin
  Camarasa, Nish Chaturvedi, et~al.
\newblock Where is {VALDO}? vascular lesions detection and segmentation
  challenge at {MICCAI} 2021.
\newblock \emph{Medical Image Analysis}, 91:\penalty0 103029, 2024.

\bibitem[Wald et~al.(2025)Wald, Roy, Isensee, Ulrich, Ziegler, Trofimova,
  Stock, Baumgartner, K{\"o}hler, and Maier-Hein]{primus}
Tassilo Wald, Saikat Roy, Fabian Isensee, Constantin Ulrich, Sebastian Ziegler,
  Dasha Trofimova, Raphael Stock, Michael Baumgartner, Gregor K{\"o}hler, and
  Klaus Maier-Hein.
\newblock Primus: Enforcing attention usage for 3d medical image segmentation.
\newblock \emph{arXiv preprint arXiv:2503.01835}, 2025.

\bibitem[Wang and Dai(2025)]{rankseg}
Zixun Wang and Ben Dai.
\newblock {RankSEG-RMA}: An efficient segmentation algorithm via reciprocal
  moment approximation.
\newblock In \emph{Advances in Neural Information Processing Systems}, 2025.

\end{thebibliography}
}

\clearpage
\appendix
\twocolumn[{\centering\Large\textbf{Supplementary Material}\par\vspace{1.0em}}]

\makeatletter
\setlength{\@fptop}{0pt}
\setlength{\@fpsep}{12pt plus 2pt}
\setlength{\@fpbot}{0pt plus 1fil}
\makeatother

\noindent Secs.~\ref{sec:supp-global-gradients}--\ref{sec:supp-union-gradients}
give the gradient derivations, and
Secs.~\ref{sec:supp-sgd}--\ref{sec:data-use} give further experimental
details. We use the notation from Secs.~\ref{sec:preliminaries}
and~\ref{sec:volume-bias} throughout.

\section{Gradient totals for global losses}
\label{sec:supp-global-gradients}

We derive Eq.~\ref{eq:global-component-gradients} for the
controlled cases in Sec.~\ref{sec:volume-bias}.

\subsection{Global cross-entropy}

Differentiating the patch-mean CE gives
\begin{equation}
 \frac{\partial\mathcal L_{\mathrm{CE}}(\Omega)}{\partial p_i}
 =\frac{1}{|\Omega|}
   \left(-\frac{y_i}{p_i}+\frac{1-y_i}{1-p_i}\right).
 \label{eq:supp-global-ce-derivative}
\end{equation}
On a missed lesion, $y_i=1$ and $p_i=1-r$, so the derivative is
$-1/(|\Omega|(1-r))$. On a false positive, $y_i=0$ and $p_i=r$,
so it is $1/(|\Omega|(1-r))$. The absolute derivative is therefore
the same at every error voxel in both cases. Summing over each component gives
\begin{equation}
 \begin{aligned}
  G_{\mathrm{CE}}(C_j)&=\frac{s_j}{|\Omega|(1-r)},\\
  G_{\mathrm{CE}}(F_l)&=\frac{f_l}{|\Omega|(1-r)}.
 \end{aligned}
 \label{eq:supp-global-ce-totals}
\end{equation}

\subsection{Global Dice}

Write $P=\sum_{i\in\Omega}p_i$ for the predicted foreground mass and
$I=\sum_{i\in\Omega}p_i y_i$ for the soft overlap. Then
\begin{equation}
 \mathcal L_{\mathrm{Dice}}(\Omega)=1-\frac{2I}{P+T}.
\end{equation}
Since $\partial I/\partial p_i=y_i$ and $\partial P/\partial p_i=1$,
\begin{equation}
 \frac{\partial\mathcal L_{\mathrm{Dice}}(\Omega)}{\partial p_i}
 =\frac{2I-2y_i(P+T)}{(P+T)^2}.
 \label{eq:supp-global-dice-derivative}
\end{equation}

\noindent\textbf{Missed lesion.}
The correctly predicted lesions contribute $T-s_j$ to both $P$ and $I$.
The missed lesion contributes $(1-r)s_j$, while the background contributes
zero. Thus
\begin{equation}
 P=I=T-s_j+(1-r)s_j=T-rs_j.
\end{equation}
For each $i\in C_j$, $y_i=1$, so
\begin{equation}
 \begin{aligned}
  \frac{\partial\mathcal L_{\mathrm{Dice}}(\Omega)}{\partial p_i}
   &=\frac{2(T-rs_j)-2(2T-rs_j)}{(2T-rs_j)^2}\\
   &=-\frac{2T}{(2T-rs_j)^2}.
 \end{aligned}
\end{equation}
Summing the absolute derivative over the $s_j$ voxels gives
\begin{equation}
 G_{\mathrm{Dice}}(C_j)=\frac{2Ts_j}{(2T-rs_j)^2}.
 \label{eq:supp-global-dice-fn-total}
\end{equation}

\medskip\noindent\textbf{False-positive component.}
All ground-truth lesions are predicted perfectly, contributing $T$ to both
$P$ and $I$. The false-positive component adds $rf_l$ to $P$ but nothing
to $I$. Hence
\begin{equation}
 P=T+rf_l,\qquad I=T.
\end{equation}
For each $i\in F_l$, $y_i=0$, giving
\begin{equation}
 \frac{\partial\mathcal L_{\mathrm{Dice}}(\Omega)}{\partial p_i}
 =\frac{2T}{(2T+rf_l)^2}.
\end{equation}
Summing over the $f_l$ voxels gives
\begin{equation}
 G_{\mathrm{Dice}}(F_l)=\frac{2Tf_l}{(2T+rf_l)^2}.
 \label{eq:supp-global-dice-fp-total}
\end{equation}

\section{Gradient totals for CC and blob}
\label{sec:supp-component-gradients}

We derive Eqs.~\ref{eq:reference-component-gradients}
and~\ref{eq:fp-component-gradients} using the same controlled cases as in
Sec.~\ref{sec:supp-global-gradients}.

\subsection{Component terms}

For component $C_j$, CC computes its Dice term over $W_j=V_j$. Blob
computes it over $W_j=C_j\cup(\Omega\setminus Y)$, which covers $C_j$
and all background voxels. In both cases,
the only ground-truth lesion in $W_j$ is $C_j$. Write
\begin{equation}
 P_j=\sum_{i\in W_j}p_i,\qquad I_j=\sum_{i\in C_j}p_i.
\end{equation}
Each Dice term and its contribution to the regional loss are then
\begin{equation}
 D_j=1-\frac{2I_j}{P_j+s_j},\qquad
 \mathcal L=\frac{1}{N}\sum_{j=1}^N D_j.
 \label{eq:supp-component-dice-loss}
\end{equation}
The voxel derivative follows from
Eq.~\ref{eq:supp-global-dice-derivative}, replacing $P$, $I$, and $T$
with $P_j$, $I_j$, and $s_j$. Voxels outside $W_j$ have zero derivative
for that term.

\subsection{Missed lesion}

For the missed lesion $C_j$, only the $j$th term depends on its
probabilities: (a) CC assigns $C_j$ to $V_j$ and (b) blob masks $C_j$
in every other term. In $W_j$, the ground-truth volume is $T_j=s_j$.
The background is predicted perfectly, so
\begin{equation}
 P_j=I_j=(1-r)s_j,\qquad P_j+T_j=(2-r)s_j.
\end{equation}
For each $i\in C_j$, the derivative of the averaged loss is
\begin{equation}
 \begin{aligned}
  \frac{\partial\mathcal L}{\partial p_i}
   &=\frac{1}{N}
     \frac{2(1-r)s_j-2(2-r)s_j}{((2-r)s_j)^2}\\
   &=-\frac{2s_j}{N(2-r)^2s_j^2}.
 \end{aligned}
\end{equation}
Summing the absolute derivative over $C_j$ gives
\begin{equation}
 \begin{aligned}
  G_{\mathrm{CC}}(C_j)=G_{\mathrm{blob}}(C_j)
   &=\frac{2s_j^2}{N(2-r)^2s_j^2}\\
   &=\frac{2}{N(2-r)^2}.
 \end{aligned}
\end{equation}
The lesion size therefore cancels at fixed confidence.

\subsection{False-positive component}

Assume $F_l\subseteq V_j\setminus Y$, so the entire false-positive
component belongs to one CC cell. All ground-truth lesions are predicted
perfectly.

\noindent\textbf{CC.}
Only the $j$th term includes $F_l$. Its predicted mass and overlap are
\begin{equation}
 P_j=s_j+rf_l,\qquad I_j=s_j.
\end{equation}
Since $y_i=0$ for each $i\in F_l$, the voxel derivative is
\begin{equation}
 \frac{\partial\mathcal L_{\mathrm{CC}}}{\partial p_i}
 =\frac{1}{N}\frac{2s_j}{(2s_j+rf_l)^2}.
\end{equation}
Summing over the $f_l$ voxels gives
\begin{equation}
 G_{\mathrm{CC}}(F_l)
 =\frac{f_l}{N}\frac{2s_j}{(2s_j+rf_l)^2}.
\end{equation}

\medskip\noindent\textbf{Example.} Consider a 50-voxel false positive at confidence
$r=0.8$ in a patch with two ground-truth lesions of 100 and 2,000 voxels.
If the false positive lies entirely in the smaller lesion's cell, its CC
gradient total is $0.087$. In the larger lesion's cell, the total is
$0.0061$, about 14 times smaller.

\medskip\noindent\textbf{Blob.}
Every blob term includes $F_l$ because $F_l$ lies in the background. For the
$k$th term ($k=1,\dots,N$),
\begin{equation}
 P_k=s_k+rf_l,\qquad I_k=s_k.
\end{equation}
Each term has a positive derivative on $F_l$, so their contributions add:
\begin{equation}
 \frac{\partial\mathcal L_{\mathrm{blob}}}{\partial p_i}
 =\frac{1}{N}\sum_{k=1}^N\frac{2s_k}{(2s_k+rf_l)^2},
 \qquad i\in F_l.
\end{equation}
Summing over the component gives
\begin{equation}
 G_{\mathrm{blob}}(F_l)
 =\frac{f_l}{N}\sum_{k=1}^N\frac{2s_k}{(2s_k+rf_l)^2}.
\end{equation}

\medskip\noindent\textbf{Example.} Consider a 50-voxel false positive at confidence
$r=0.8$ near a correctly predicted 100-voxel lesion. With this single
ground-truth lesion, the false positive's blob gradient total is $0.174$.
Adding a separate, correctly predicted 20-voxel lesion anywhere in the patch
raises the total to $0.243$. Increasing the added lesion's size to 2,000
voxels lowers the total to $0.093$.

For both losses, the total gradient depends on the size of the false positive.
CC weights the component through its assigned lesion, while blob includes a
contribution from every ground-truth lesion.

\subsection{Conversion to logit gradients}

For $p_i=\operatorname{sigmoid}(z_i)$, the chain rule gives
\begin{equation}
 \frac{\partial\mathcal L}{\partial z_i}
 =\frac{\partial\mathcal L}{\partial p_i}p_i(1-p_i).
\end{equation}
In either controlled case, $p_i(1-p_i)=r(1-r)$ throughout the error
region $A\in\{C_j,F_l\}$. Therefore,
\begin{equation}
 \begin{aligned}
  G^z_{\mathcal L}(A)
   &=\sum_{i\in A}\left|\frac{\partial\mathcal L}{\partial z_i}\right|\\
   &=r(1-r)\sum_{i\in A}
     \left|\frac{\partial\mathcal L}{\partial p_i}\right|\\
   &=r(1-r)G_{\mathcal L}(A).
 \end{aligned}
 \label{eq:supp-controlled-logit-totals}
\end{equation}
This applies to both global and component-based losses above. At fixed
confidence, the common factor preserves the size comparisons.

\subsection{Cross-entropy terms}
\label{sec:supp-component-ce}

Blob averages each masked CE term over the whole patch. Each foreground
voxel appears in one term, while each background voxel appears in all
$N$ terms. Averaging over the components therefore gives
\begin{equation}
 \mathcal L_{\mathrm{blob}}^{\mathrm{CE}}
 =\frac{1}{N|\Omega|}\sum_{i\in Y}\ell_i
  +\frac{1}{|\Omega|}\sum_{i\in\Omega\setminus Y}\ell_i.
 \label{eq:supp-blob-ce}
\end{equation}
Each background voxel thus receives $N$ times the weight of each
foreground voxel. For $N=1$, this reduces to global CE.

CC averages CE within each Voronoi cell, then averages over the cells:
\begin{equation}
 \mathcal L_{\mathrm{CC}}^{\mathrm{CE}}
 =\frac{1}{N}\sum_{j=1}^N\frac{1}{|V_j|}
   \sum_{i\in V_j}\ell_i.
 \label{eq:supp-cc-ce}
\end{equation}
Every voxel in $V_j$ receives weight $1/(N|V_j|)$.
Neither CE term explicitly normalizes false positives by their own size.

\section{Dice gradient failure modes}
\label{sec:supp-ce-dice-gradients}

We derive the gradient properties in Sec.~\ref{sec:why-ce} and the
background-to-foreground gradient ratio in Eq.~\ref{eq:gt-gradient-balance}.
We hold the cell assignments and weights fixed when differentiating.

\subsection{Weighted cross-entropy}

Let $\mathcal L=\sum_j a_j\ell_j$, where $a_j\geq0$ are fixed weights.
Since each $\ell_j$ depends only on $p_j=\operatorname{sigmoid}(z_j)$,
\begin{equation}
 \begin{aligned}
  \frac{\partial\mathcal L}{\partial z_i}
   &=a_i\left(-\frac{y_i}{p_i}+\frac{1-y_i}{1-p_i}\right)
     p_i(1-p_i)\\
   &=a_i\bigl[-y_i(1-p_i)+(1-y_i)p_i\bigr]\\
   &=a_i(p_i-y_i).
 \end{aligned}
 \label{eq:supp-weighted-ce-logit-gradient}
\end{equation}
Thus the weight $a_i$ directly scales the voxel's logit gradient.
Exchanging foreground and background sends $(p_i,y_i)$ to
$(1-p_i,1-y_i)$ and changes only the sign of this gradient.

\subsection{Confident errors}
\label{sec:supp-confident-errors}

For a foreground error with $p_i=1-r$ or a background error with $p_i=r$,
the weighted CE logit gradient satisfies
\begin{equation}
 \left|\frac{\partial\mathcal L_{\mathrm{CE}}}{\partial z_i}\right|
 =a_i r\ \longrightarrow\ a_i
 \qquad\text{as }r\to1.
\end{equation}
For unsmoothed Dice in a region $V$ with $T>0$,
Eq.~\ref{eq:supp-global-dice-derivative} gives
\begin{equation}
 \frac{\partial\mathcal L_{\mathrm{Dice}}(V)}{\partial z_i}
 =\frac{2I-2y_i(P+T)}{(P+T)^2}\,r(1-r)
 \longrightarrow 0
 \qquad\text{as }r\to1.
 \label{eq:supp-confident-dice-limit}
\end{equation}

\medskip\noindent\textbf{Example.} In the missed-lesion case of
Eq.~\ref{eq:supp-global-dice-fn-total}, let $s_j=10$ voxels, or $10\%$ of
the $T=100$ foreground voxels. As error confidence $r$ rises from $0.90$ to
$0.99$, the lesion's Dice gradient total drops from
$4.9\times10^{-3}$ to $5.5\times10^{-4}$. With a common voxel weight $a$,
the corresponding CE total rises from $9.0a$ to $9.9a$.

\subsection{Background and foreground weighting}

Consider a foreground error and a background error in the same cell,
with $p_{\mathrm{fg}}=1-r$ and $p_{\mathrm{bg}}=r$.
Write $d=2I/(P+T)$ for the cell's unsmoothed Dice score.
By Eq.~\ref{eq:supp-confident-dice-limit}, their absolute logit gradients are
\begin{equation}
 \begin{aligned}
  \left|\frac{\partial\mathcal L_{\mathrm{Dice}}(V)}
                   {\partial z_{\mathrm{fg}}}\right|
   &=\frac{2-d}{P+T}r(1-r),\\
  \left|\frac{\partial\mathcal L_{\mathrm{Dice}}(V)}
                   {\partial z_{\mathrm{bg}}}\right|
   &=\frac{d}{P+T}r(1-r).
 \end{aligned}
\end{equation}
Taking their ratio gives
\begin{equation}
 \frac{|\partial\mathcal L_{\mathrm{Dice}}/\partial z_{\mathrm{bg}}|}
      {|\partial\mathcal L_{\mathrm{Dice}}/\partial z_{\mathrm{fg}}|}
 =\frac{d}{2-d}\leq1.
\end{equation}
For CE with a common weight $a>0$ within the cell, both errors instead
have absolute gradient $ar$, giving a ratio of one. Dice therefore gives
the background error less relative weight when $d<1$, with the ratio
approaching zero as $d\to0$.

\subsection{Dice with smoothing}
\label{sec:supp-dice-smoothing}

For smoothed Dice, the denominator is $S=P+T+\epsilon$.
Differentiating the loss in Eq.~\ref{eq:base-losses} gives
\begin{equation}
 \frac{\partial\mathcal L_{\mathrm{Dice}}(V)}{\partial p_i}
 =\frac{2I+\epsilon-2y_i S}{S^2}.
 \label{eq:supp-smoothed-dice-derivative}
\end{equation}
In a cell without ground-truth lesions, $T=I=0$ and the loss is
$1-\epsilon/(P+\epsilon)$. For $P+\epsilon>0$, the gradients are
\begin{equation}
 \begin{aligned}
  \frac{\partial\mathcal L_{\mathrm{Dice}}(V)}{\partial p_i}
   &=\frac{\epsilon}{(P+\epsilon)^2},\\
  \frac{\partial\mathcal L_{\mathrm{Dice}}(V)}{\partial z_i}
   &=\frac{\epsilon p_i(1-p_i)}{(P+\epsilon)^2}.
 \end{aligned}
 \label{eq:supp-empty-dice-gradients}
\end{equation}
With $\epsilon=0$ and $P>0$, the gradient is zero.

For $\epsilon>0$, consider a false-positive component $C$ of $n$ voxels
with $p_i=p$ throughout $C$. Let $P_{\mathrm{out}}=\sum_{i\in V\setminus C}p_i$
be the predicted mass elsewhere in the cell. Substituting
$P=np+P_{\mathrm{out}}$ into Eq.~\ref{eq:supp-empty-dice-gradients} and summing
over $C$ gives
\begin{equation}
 \sum_{i\in C}\left|\frac{\partial\mathcal L_{\mathrm{Dice}}(V)}{\partial z_i}\right|
 =\frac{n\epsilon p(1-p)}{(np+P_{\mathrm{out}}+\epsilon)^2}.
\end{equation}
With $P_{\mathrm{out}}\ll np+\epsilon$, the denominator is approximately
$(np+\epsilon)^2$, giving Eq.~\ref{eq:smoothed-fp-pull}. At fixed $p$, this
approximation peaks at $n=\epsilon/p$ when $n$ is treated as continuous.

\medskip\noindent\textbf{Example.} With $\epsilon=10$ and $p=0.8$, the Dice gradient
totals for components of $1$, $13$, and $100$ voxels are $0.014$, $0.050$,
and $0.020$, respectively.

\section{Gradient scaling in UnionCE}
\label{sec:supp-union-gradients}

We derive the effect of the size normalizer in
Eq.~\ref{eq:cluster-normalizer} on the regional objective in
Eq.~\ref{eq:union-ce}. The prediction mask, union components, Voronoi
cells, and normalizers are held fixed when differentiating.
We assume $K\geq1$. Each cell contains its union component
$C_k=U\cap V_k$, so at least one of $g_k$ and $h_k$ is positive and
$s_k\geq1$.

\subsection{Component gradients}

Each voxel belongs to exactly one cell. For $i\in V_k$, its weight in
the regional CE is $1/(Ks_k)$. Applying
Eq.~\ref{eq:supp-weighted-ce-logit-gradient} gives
\begin{equation}
 \frac{\partial\mathcal L_{\mathrm{union}}}{\partial z_i}
 =\frac{p_i-y_i}{Ks_k}.
 \label{eq:supp-union-logit-gradient}
\end{equation}
For any set $A\subseteq V_k$, the total absolute logit gradient is therefore
\begin{equation}
 G^z_{\mathrm{union}}(A)
 =\frac{1}{Ks_k}\sum_{i\in A}|p_i-y_i|.
 \label{eq:supp-union-region-total}
\end{equation}

\subsection{Isolated errors}

Consider a union component of $n$ voxels representing either
(a) a missed lesion, with $g_k=n$, $h_k=0$, and $p_i=1-r$ on $C_k$,
or (b) an isolated false positive, with $g_k=0$, $h_k=n$, and $p_i=r$
on $C_k$. In both cases, $s_k=n$ and $|p_i-y_i|=r$ throughout the
component. Hence
\begin{equation}
 G^z_{\mathrm{union}}(C_k)
 =\frac{nr}{Kn}=\frac{r}{K}.
 \label{eq:supp-union-isolated-total}
\end{equation}
At fixed confidence and component count $K$, this total is independent
of the component size. This cancellation holds regardless of the
probabilities on surrounding background voxels within the cell.

\subsection{Surrounding background}

Let $B_k=V_k\setminus C_k$. These voxels belong to neither the
ground-truth nor the prediction mask, so $y_i=0$ on $B_k$. Their total
regional logit gradient is
\begin{equation}
 G^z_{\mathrm{union}}(B_k)=\frac{1}{Ks_k}\sum_{i\in B_k}p_i.
\end{equation}
The total over the entire cell is therefore
\begin{equation}
 G^z_{\mathrm{union}}(V_k)
 =G^z_{\mathrm{union}}(C_k)
  +\frac{1}{Ks_k}\sum_{i\in B_k}p_i.
\end{equation}
Thus equal component totals can coexist with different cell totals,
depending on the predicted foreground mass in the surrounding background.

\section{SGD training recipe}
\label{sec:supp-sgd}

Table~\ref{tab:sgd} reports results for the three baselines trained with
nnU-Net's SGD recipe (ablation~6). The \method{} row gives the AdamW
results from Table~\ref{tab:main} for comparison.

\begin{table}[t]
\caption{U-Net results with nnU-Net's SGD recipe for the baselines (ablation~6).
\method is as in Table~\ref{tab:main}.
Tests vs. \method, Holm over three baselines.}
\label{tab:sgd}
\centering\footnotesize
\setlength{\tabcolsep}{1pt}
\renewcommand{\arraystretch}{1.1}
\begin{tabular*}{\linewidth}{@{\extracolsep{\fill}}lrrrrrr@{}}
\toprule
Method & Dice $\uparrow$ & C-DSC $\uparrow$ & F1 $\uparrow$ & Prec. $\uparrow$ & Rec. $\uparrow$ & FP/neg $\downarrow$\\
\midrule
\multicolumn{7}{l}{\textit{AutoPET III}}\\
DiceCE & 56.5\textsuperscript{$\dagger$} & \underline{32.7}\textsuperscript{$\dagger$} & \underline{53.0}\textsuperscript{$\dagger$} & \underline{50.0}\textsuperscript{$\dagger$} & 66.3\textsuperscript{$\dagger$} & \underline{5.5}\textsuperscript{$\dagger$}\\
blob loss & 54.5\textsuperscript{$\dagger$} & 30.9\textsuperscript{$\dagger$} & 52.1\textsuperscript{$\dagger$} & 47.3\textsuperscript{$\dagger$} & 68.8\textsuperscript{$\dagger$} & 6.7\textsuperscript{$\dagger$}\\
CC-DiceCE & \underline{57.2}\textsuperscript{$\dagger$} & 26.7\textsuperscript{$\dagger$} & 46.2\textsuperscript{$\dagger$} & 37.6\textsuperscript{$\dagger$} & \textbf{73.7}\textsuperscript{$\dagger$} & 10.8\textsuperscript{$\dagger$}\\
\textbf{\method} & \textbf{61.8} & \textbf{42.1} & \textbf{62.8} & \textbf{63.4} & \underline{69.7} & \textbf{3.2}\\
\midrule
\multicolumn{7}{l}{\textit{ISLES}}\\
DiceCE & \textbf{65.2}\textsuperscript{$\dagger$} & \underline{41.2} & \underline{59.1}\textsuperscript{$\dagger$} & \underline{65.9}\textsuperscript{$\dagger$} & 64.6 & 3.0\\
blob loss & 64.4\textsuperscript{$\dagger$} & 38.7\textsuperscript{$\dagger$} & 57.1\textsuperscript{$\dagger$} & 60.2\textsuperscript{$\dagger$} & \underline{67.0}\textsuperscript{$\dagger$} & \textbf{1.8}\\
CC-DiceCE & \underline{65.0}\textsuperscript{$\dagger$} & 21.0\textsuperscript{$\dagger$} & 36.0\textsuperscript{$\dagger$} & 29.7\textsuperscript{$\dagger$} & \textbf{71.6}\textsuperscript{$\dagger$} & \underline{2.8}\\
\textbf{\method} & 64.4 & \textbf{42.8} & \textbf{62.4} & \textbf{72.9} & 64.7 & \underline{2.8}\\
\midrule
\multicolumn{7}{l}{\textit{BraTS-METS}}\\
DiceCE & 74.8\textsuperscript{$\dagger$} & \underline{55.1}\textsuperscript{$\dagger$} & \underline{73.8}\textsuperscript{$\dagger$} & \underline{80.5}\textsuperscript{$\dagger$} & 73.7\textsuperscript{$\dagger$} & \underline{1.2}\\
blob loss & 71.8\textsuperscript{$\dagger$} & 52.5\textsuperscript{$\dagger$} & 73.2\textsuperscript{$\dagger$} & 78.6\textsuperscript{$\dagger$} & 74.3\textsuperscript{$\dagger$} & \textbf{1.1}\\
CC-DiceCE & \underline{75.3}\textsuperscript{$\dagger$} & 49.8\textsuperscript{$\dagger$} & 69.7\textsuperscript{$\dagger$} & 68.0\textsuperscript{$\dagger$} & \textbf{79.4}\textsuperscript{$\dagger$} & 1.3\\
\textbf{\method} & \textbf{76.6} & \textbf{59.1} & \textbf{77.4} & \textbf{83.5} & \underline{77.1} & \textbf{1.1}\\
\midrule
\multicolumn{7}{l}{\textit{LiTS}}\\
DiceCE & 57.4 & \underline{31.4}\textsuperscript{$\dagger$} & \underline{52.3}\textsuperscript{$\dagger$} & \underline{47.0}\textsuperscript{$\dagger$} & 71.1 & \underline{3.8}\textsuperscript{$\dagger$}\\
blob loss & 54.4\textsuperscript{$\dagger$} & 30.0\textsuperscript{$\dagger$} & 51.3\textsuperscript{$\dagger$} & 44.4\textsuperscript{$\dagger$} & 74.4 & 4.5\textsuperscript{$\dagger$}\\
CC-DiceCE & \underline{58.9} & 28.6\textsuperscript{$\dagger$} & 49.5\textsuperscript{$\dagger$} & 40.0\textsuperscript{$\dagger$} & \textbf{79.2}\textsuperscript{$\dagger$} & 4.2\textsuperscript{$\dagger$}\\
\textbf{\method} & \textbf{60.7} & \textbf{45.8} & \textbf{69.3} & \textbf{69.4} & \underline{75.4} & \textbf{1.6}\\
\midrule
\multicolumn{7}{l}{\textit{CMB}}\\
DiceCE & \underline{44.8} & 37.7 & \underline{58.4} & 62.0 & \underline{62.4} & \underline{1.1}\textsuperscript{$\dagger$}\\
blob loss & 43.3 & 36.4 & 55.9 & 59.9 & 60.5 & 1.4\textsuperscript{$\dagger$}\\
CC-DiceCE & \textbf{47.8} & \textbf{40.9} & \textbf{61.8} & \underline{64.3} & \textbf{65.3}\textsuperscript{$\dagger$} & \underline{1.1}\textsuperscript{$\dagger$}\\
\textbf{\method} & 44.0 & \underline{38.8} & 57.6 & \textbf{65.0} & 57.6 & \textbf{0.5}\\
\bottomrule
\end{tabular*}
\end{table}

\section{Inference with fixed checkpoints}

Table~\ref{tab:inference} compares the two inference setups on the same
checkpoints (ablation~7). Each entry gives the score with \method{}
inference minus the score with logit averaging and argmax
(nnU-Net's default).

\begin{table}[t]
\caption{Inference swap (ablation~7). Change from baseline inference (logit averaging and argmax) to
\method inference (probability averaging, top-$k$ selection, and $\tau=0.1$)
for the same U-Net checkpoints (\method minus baseline inference). Scores change in
percentage points, FP/neg in counts. Holm over four swaps.}
\label{tab:inference}
\centering\footnotesize
\setlength{\tabcolsep}{2pt}
\renewcommand{\arraystretch}{0.95}
\begin{tabular*}{\columnwidth}{@{\extracolsep{\fill}}lrrrrrr@{}}
\toprule
Loss & Dice & C-DSC & F1 & Prec. & Rec. & FP/neg\\
\midrule
\multicolumn{7}{l}{\textit{AutoPET III}}\\
DiceCE & $-$1.0\textsuperscript{$\dagger$} & $-$1.2\textsuperscript{$\dagger$} & $-$1.6\textsuperscript{$\dagger$} & $-$2.5\textsuperscript{$\dagger$} & +1.1\textsuperscript{$\dagger$} & +1.3\textsuperscript{$\dagger$}\\
blob loss & 0.0\textsuperscript{$\dagger$} & $-$0.3 & $-$1.5\textsuperscript{$\dagger$} & $-$2.1\textsuperscript{$\dagger$} & +1.2\textsuperscript{$\dagger$} & +1.5\textsuperscript{$\dagger$}\\
CC-DiceCE & $-$1.0\textsuperscript{$\dagger$} & $-$1.7\textsuperscript{$\dagger$} & $-$2.6\textsuperscript{$\dagger$} & $-$3.1\textsuperscript{$\dagger$} & +1.1\textsuperscript{$\dagger$} & +2.9\textsuperscript{$\dagger$}\\
\method & +7.4\textsuperscript{$\dagger$} & +5.9\textsuperscript{$\dagger$} & +4.8\textsuperscript{$\dagger$} & +0.9 & +9.8\textsuperscript{$\dagger$} & +1.2\textsuperscript{$\dagger$}\\
\midrule
\multicolumn{7}{l}{\textit{ISLES}}\\
DiceCE & +0.3 & $-$0.1\textsuperscript{$\dagger$} & $-$0.1 & $-$0.8\textsuperscript{$\dagger$} & +1.1\textsuperscript{$\dagger$} & +2.2\\
blob loss & +0.6\textsuperscript{$\dagger$} & $-$0.5 & 0.0\textsuperscript{$\dagger$} & $-$1.3\textsuperscript{$\dagger$} & +1.9\textsuperscript{$\dagger$} & +1.2\\
CC-DiceCE & +0.7 & $-$0.6\textsuperscript{$\dagger$} & $-$0.7\textsuperscript{$\dagger$} & $-$1.2\textsuperscript{$\dagger$} & +1.4\textsuperscript{$\dagger$} & +2.6\\
\method & +3.2\textsuperscript{$\dagger$} & +1.4\textsuperscript{$\dagger$} & +2.0\textsuperscript{$\dagger$} & $-$0.5 & +4.7\textsuperscript{$\dagger$} & +1.8\\
\midrule
\multicolumn{7}{l}{\textit{BraTS-METS}}\\
DiceCE & 0.0 & +0.2\textsuperscript{$\dagger$} & +0.3 & $-$0.6 & +1.3\textsuperscript{$\dagger$} & +1.5\textsuperscript{$\dagger$}\\
blob loss & +0.6\textsuperscript{$\dagger$} & +0.3\textsuperscript{$\dagger$} & +0.1 & $-$0.9\textsuperscript{$\dagger$} & +1.4\textsuperscript{$\dagger$} & +1.5\textsuperscript{$\dagger$}\\
CC-DiceCE & +0.5\textsuperscript{$\dagger$} & $-$0.8 & $-$0.4\textsuperscript{$\dagger$} & $-$1.5\textsuperscript{$\dagger$} & +1.5\textsuperscript{$\dagger$} & +2.5\textsuperscript{$\dagger$}\\
\method & +3.3\textsuperscript{$\dagger$} & +3.2\textsuperscript{$\dagger$} & +2.2\textsuperscript{$\dagger$} & +0.8 & +3.6\textsuperscript{$\dagger$} & +0.7\textsuperscript{$\dagger$}\\
\midrule
\multicolumn{7}{l}{\textit{LiTS}}\\
DiceCE & +0.4 & $-$2.4 & $-$1.7\textsuperscript{$\dagger$} & $-$4.3\textsuperscript{$\dagger$} & +4.8\textsuperscript{$\dagger$} & +1.9\textsuperscript{$\dagger$}\\
blob loss & $-$0.2 & $-$2.6\textsuperscript{$\dagger$} & $-$3.6\textsuperscript{$\dagger$} & $-$4.8\textsuperscript{$\dagger$} & +0.9 & +3.4\textsuperscript{$\dagger$}\\
CC-DiceCE & +0.1 & $-$4.9\textsuperscript{$\dagger$} & $-$6.6\textsuperscript{$\dagger$} & $-$8.6\textsuperscript{$\dagger$} & +2.2\textsuperscript{$\dagger$} & +3.8\textsuperscript{$\dagger$}\\
\method & +1.9 & +2.0\textsuperscript{$\dagger$} & +2.2 & $-$0.1 & +5.1\textsuperscript{$\dagger$} & +0.9\textsuperscript{$\dagger$}\\
\midrule
\multicolumn{7}{l}{\textit{CMB}}\\
DiceCE & +3.8\textsuperscript{$\dagger$} & +2.4 & +2.4 & +0.9 & +4.9\textsuperscript{$\dagger$} & +0.5\textsuperscript{$\dagger$}\\
blob loss & +5.7\textsuperscript{$\dagger$} & +4.4\textsuperscript{$\dagger$} & +5.3 & +4.1 & +6.8\textsuperscript{$\dagger$} & +0.7\textsuperscript{$\dagger$}\\
CC-DiceCE & +5.1\textsuperscript{$\dagger$} & +3.9\textsuperscript{$\dagger$} & +4.2 & +4.5 & +5.5\textsuperscript{$\dagger$} & +0.7\textsuperscript{$\dagger$}\\
\method & +8.1\textsuperscript{$\dagger$} & +7.4\textsuperscript{$\dagger$} & +9.2\textsuperscript{$\dagger$} & +9.7\textsuperscript{$\dagger$} & +10.0\textsuperscript{$\dagger$} & +0.2\textsuperscript{$\dagger$}\\
\bottomrule
\end{tabular*}
\end{table}

\section{Global cross-entropy}

Table~\ref{tab:ce} compares global CE with DiceCE (ablation~10).
The global CE variant keeps \method's training recipe and inference
while dropping the regional loss term.

\begin{table}[t]
\caption{Global CE and DiceCE with U-Net (ablation~10). DiceCE uses
argmax. CE uses top-$k$. Bold marks the best displayed
value. Daggers mark uncorrected CE versus DiceCE tests.}
\label{tab:ce}
\centering\footnotesize
\setlength{\tabcolsep}{2pt}
\renewcommand{\arraystretch}{0.95}
\begin{tabular*}{\columnwidth}{@{\extracolsep{\fill}}lrrrrrr@{}}
\toprule
Loss & Dice $\uparrow$ & C-DSC $\uparrow$ & F1 $\uparrow$ & Prec. $\uparrow$ & Rec. $\uparrow$ & FP/neg $\downarrow$\\
\midrule
\multicolumn{7}{l}{\textit{AutoPET III}}\\
DiceCE & \textbf{59.3} & \textbf{35.1} & \textbf{55.2} & \textbf{52.4} & 67.1 & \textbf{5.1}\\
CE & 57.8 & 33.3\textsuperscript{$\dagger$} & 53.3\textsuperscript{$\dagger$} & 49.9\textsuperscript{$\dagger$} & \textbf{71.1}\textsuperscript{$\dagger$} & 10.7\textsuperscript{$\dagger$}\\
\midrule
\multicolumn{7}{l}{\textit{ISLES}}\\
DiceCE & \textbf{65.5} & \textbf{41.6} & \textbf{59.2} & \textbf{65.4} & 65.6 & \textbf{1.4}\\
CE & 64.5\textsuperscript{$\dagger$} & 36.8\textsuperscript{$\dagger$} & 53.8\textsuperscript{$\dagger$} & 56.5\textsuperscript{$\dagger$} & \textbf{66.2} & 4.6\\
\midrule
\multicolumn{7}{l}{\textit{BraTS-METS}}\\
DiceCE & 75.3 & 55.7 & 74.1 & \textbf{80.3} & 74.6 & \textbf{1.2}\\
CE & \textbf{76.6}\textsuperscript{$\dagger$} & \textbf{56.6} & \textbf{74.9} & 80.1 & \textbf{76.1}\textsuperscript{$\dagger$} & 4.8\textsuperscript{$\dagger$}\\
\midrule
\multicolumn{7}{l}{\textit{LiTS}}\\
DiceCE & \textbf{57.4} & \textbf{37.1} & \textbf{58.2} & \textbf{54.9} & \textbf{72.7} & \textbf{2.8}\\
CE & 53.3\textsuperscript{$\dagger$} & 29.1\textsuperscript{$\dagger$} & 48.2\textsuperscript{$\dagger$} & 43.9\textsuperscript{$\dagger$} & 69.9 & 15.4\textsuperscript{$\dagger$}\\
\midrule
\multicolumn{7}{l}{\textit{CMB}}\\
DiceCE & 43.9 & 37.6 & 56.5 & 61.1 & 59.2 & \textbf{0.9}\\
CE & \textbf{48.2} & \textbf{41.2} & \textbf{62.7} & \textbf{68.9} & \textbf{65.4} & 1.0\\
\bottomrule
\end{tabular*}
\end{table}

\begin{table}[t]
\caption{Training time per epoch on BraTS-METS. Each loss trains for
10 epochs from its trained fold-0 checkpoint. We take the median epoch time of each run and report the mean over three runs.}
\label{tab:runtime}
\centering\footnotesize
\setlength{\tabcolsep}{4pt}
\begin{tabular*}{\columnwidth}{@{\extracolsep{\fill}}lrrrr@{}}
\toprule
& \multicolumn{2}{c}{U-Net} & \multicolumn{2}{c}{PrimusV2-S}\\
\cmidrule(lr){2-3}\cmidrule(lr){4-5}
Loss & s/epoch & vs.\ DiceCE & s/epoch & vs.\ DiceCE\\
\midrule
DiceCE & 48.8 & -- & 42.9 & --\\
blob loss & 49.0 & +0.5\% & 43.6 & +1.6\%\\
CC-DiceCE & 49.4 & +1.2\% & 44.0 & +2.7\%\\
\method & 49.9 & +2.3\% & 43.6 & +1.6\%\\
\bottomrule
\end{tabular*}
\end{table}

\section{Training cost}
\label{sec:runtime}

Table~\ref{tab:runtime} compares the training time of the four losses on an
NVIDIA RTX A6000 GPU.
\method adds 2.3\% to the epoch time of DiceCE with U-Net and 1.6\% with
PrimusV2-S. This is similar to blob loss and CC-DiceCE, which add 0.5 to
2.7\%.

\section{Datasets}
\label{sec:data-use}

The \href{https://www.synapse.org/Synapse:syn74274097}{BraTS-METS} and
\href{https://autopet.org/autopetiii.html}{AutoPET III} datasets are provided under CC BY-NC 4.0.
\href{https://zenodo.org/records/4520773}{VALDO} is licensed under CC BY-NC-SA 4.0, while
\href{https://competitions.codalab.org/competitions/17094\#learn_the_details-terms_and_conditions}{LiTS}
is provided under CC BY-NC-ND 4.0.
The \href{https://isles-26.grand-challenge.org/}{ISLES'26} training data are subject to
\href{https://fcon_1000.projects.nitrc.org/indi/retro/atlas_agreement.html}{ATLAS Terms of Use}.

\end{document}